\documentclass[numbered,3p,sort&compress]{elsarticle} 
\usepackage{geometry}
\usepackage{dsfont}
\usepackage{amsmath}
\usepackage{cancel}	
\usepackage{amssymb}
\usepackage{amsbsy}
\usepackage{amsfonts}
\usepackage{latexsym}
\usepackage{bm}
\usepackage{verbatim}
\usepackage{mathrsfs}
\usepackage{graphicx}
\usepackage{color}
\usepackage{subfig}
\usepackage{wrapfig}
\usepackage{float}
\usepackage{algorithm2e}
\usepackage{algpseudocode}
\usepackage{setspace,bm}
\journal{the public}

\definecolor{red}{rgb}{1,0,0}								
\definecolor{blue}{rgb}{0,0,1}							

\newcommand{\jdh}[1]{}
\renewcommand{\jdh}[1]{{\color{green}{#1}}} 

\newcommand{\sg}[1]{}
\renewcommand{\sg}[1]{{\color{blue}{#1}}} 

\newcommand{\jd}[1]{}
\renewcommand{\jd}[1]{{\color{red}{#1}}} 

\newcommand{\bmath}[1]{\mathbf{\bm{#1}}}
\def \bx{\bmath{x}}
\def \bX{\bmath{X}}

\def \bd{\bmath{d}}
\def \bD{\bmath{D}}
\def \bA{\bmath{A}}
\def \bd{\bmath{d}}
\def \ba{\bmath{a}}

\begin{document}

\begin{frontmatter}


\title{Anomaly-Sensitive Dictionary Learning for Unsupervised Diagnostics\\ of Solid Media}

\author[CE]{Jeffrey M. Druce}

\author[ECE]{Jarvis D. Haupt}

\author[CE]{Stefano Gonella~\corref{cor1}}
\ead{sgonella@umn.edu}
\cortext[cor1]{Corresponding Author}

\address[CE]{Department of Civil Engineering, University of Minnesota, Minneapolis, MN, USA}
\address[ECE]{Department of Electrical and Computer Engineering, University of Minnesota, Minneapolis, MN, USA}

\begin{abstract}
\bfseries
This paper proposes a strategy for the detection and triangulation of structural anomalies in solid media. The method revolves around the construction of sparse representations of the medium's dynamic response, obtained by learning instructive dictionaries which form a suitable basis for the response data. The resulting sparse coding problem is recast as a modified dictionary learning task with additional spatial sparsity constraints enforced on the atoms of the learned dictionaries, which provides them with a prescribed spatial topology that is designed to unveil anomalous regions in the physical domain. The proposed methodology is model agnostic, i.e., it forsakes the need for a physical model and requires virtually no \emph{a priori} knowledge of the structure's material properties, as all the inferences are exclusively informed by the data through the layers of information that are available in the intrinsic salient structure of the material's dynamic response. This characteristic makes the approach powerful for anomaly identification in systems with unknown or heterogeneous property distribution, for which a  model is unsuitable or unreliable. The method is validated using both synthetically generated data and experimental data acquired using a scanning laser Doppler vibrometer.

\end{abstract}


\end{frontmatter}

\section{Introduction}

The past three decades have witnessed the advent of a variety of structural diagnostics methodologies based on guided waves~\cite{Staszewski,Rose}. The detection and triangulation principle underlying all methods based on guided waves (the pulse-echo or pitch-catch approaches~\cite{Pitch_Catch_Fu_Kuo_Chang}) can be seen as a structural analog of the \textit{radar} problem: waves are generated and received by transmitter-receiver pairs distributed over the test specimen, a signature of wave scattering is captured along each transmitter-receiver path, and the position of the defect is subsequently triangulated using data from multiple transducers. Numerous efforts have been dedicated to the construction of damage location estimators from measurements acquired by sparse arrays of sensors. Popular approaches include: statistical methods~\cite{Likelihood_Guided_Waves_Flynn}, acoustic imaging techniques~\cite{Michaels_Sparse}, singular value decomposition~\cite{SVD_Sensors_Damage_NCState,SVD_Sensors_TAMU}, spatial optimization of the sensor networks~\cite{Wang_Yuan_PZT_Locations_Damage_Detection}, and methods based on the time reversal operator~\cite{Prada94,Foroozan_localization_algorithms}.

Pitch-catch methods allow anomaly triangulation using parsimonious sensors data, which makes them ideally suited for online or \emph{in-situ} SHM applications~\cite{Kessler_Soutis_Lamb_insitu,Kirikera_Balogun_Krishnaswamy_SHM}, where it is crucial that the acquisition system is highly portable and easily deployable. On the other hand, they, like all methods based on the radar paradigm, can suffer major weaknesses when the assumptions on the ideality of the medium are relaxed - a common scenario in the context of aging materials and damage formation. These methods rely in fact on the possibility to detect the individual scattered signals, estimate with some precision the associated times of flight, and finally triangulate the scattering sources by applying some direct knowledge of the medium properties (e.g., the wave speed). These tasks are often hard to accomplish in the case of highly heterogeneous media, materials with extreme internal complexity (e.g., random microstructures) or materials experiencing severe property degradation, for which a material model is either unknown or unreliable.

Recently, the field of structural diagnostics has been flooded with methodologies originally developed within the field of machine learning (ML)~\cite{Farrar_ML}. The picture that emerges is one of a fast-growing field that is quickly incorporating inputs from parallel disciplines. Examples of ML techniques used in support of wave-based diagnostics include neural networks~\cite{Lamb_waves_delamination_Neural}, matching pursuit decomposition~\cite{Michaels_Matching_Pursuit,Das_et_al_MPD_SPIE,Das_Chattopadhyay_MoteCarlo_MPD,Mallat_MPD_Dictionaries}, support vector machines~\cite{Das_et_al_SPIE_SVM} and compressive sensing~\cite{Azimipanah_compressive_imaging}. It is worth pointing out that the majority of the existing ML approaches for diagnostics involve supervised learning techniques for feature classification, whose use may be hindered by the need for large training data sets and databases.

In parallel, another powerful class of diagnostic methodologies has stemmed from the availability of laser-based acquisition systems~\cite{Sharma-et-al_Damage-Index_AIAA_2006,Michaels_Ruzzene_Michaels_Ultrasonics_2010}. By means of a Scanning Laser Doppler Vibrometer (SLDV) it is possible to perform non-contact measurements of the velocity of points belonging to a (potentially very dense) scanning grid defined on an object's surface, which enables full spatial reconstruction of its vibration or wave response. A number of dedicated image processing techniques have been developed in conjunction with laser experiments to meet desired identification and visualization criteria; among them are methods based on space-time DFT~\cite{Chiu,Alleyne}, wavenumber-space filters~\cite{Ruzzene_SMS_2007} and Laplace filters~\cite{Sohn_Delamination_SMS_2011, Sohn_Laser_Crack_SMS_2013}. In sheer contrast with the radar triangulation approach, laser-based diagnostics promote a different paradigm where the inference is performed directly on a data-rich, spatially reconstructed response. While the acquisition of richer data poses additional challenges in terms of sensing system requirements, it opens new avenues for inference strategies with superior accuracy and robustness.

The possibility to turn the attention from analyzing sparse time history data to operating upon spatially reconstructed waveforms, coupled with the availability of algorithms to mine complex data structures, represents a significant shift in perspective for the diagnostics problem. At the core of this new approach is the notion that, from a data standpoint, a wavefield is essentially a data cube, slices of which represent snapshots of the dynamic response at different time instants. Treating an evolving wavefield as a collection of images immediately presents the opportunity to revisit the anomaly detection problem with the mindset and methodology of image processing and computer vision (CV)~\cite{Itti98, Itti01,Yan10, Shen12}. For example, the problem of detecting anomalies in the physical medium has a data equivalent in the problem of identifying atypical patterns in the data structure~\cite{Patcha07, Chandola09}. One of the essential concepts behind this approach is the notion that, in every region of the domain that is sufficiently far from a defect, the displacement time histories will exhibit some similar, but unknown, ``typical'' behavior, while the time histories recorded in spatial regions in the immediate vicinity of a defect will exhibit some (also unknown) signature of the defect that is different from the typical response observed in the bulk of the domain. The regions exhibiting atypical behavior are referred to as \emph{salient}.  When only a few regions exhibit atypical behavior, the notion of \emph{saliency} can be viewed as a generalization of the concept of \emph{sparsity}, which has played a central role in signal processing, statistics, and machine learning research in recent years; see, e.g.,~\cite{Chen_2001_ADB_588736_588850, Candes1, Donoho, Candes2, HauptRP, tropp2004greed, tropp2006just, Bruckstein}. In the context of wave-based structural diagnostics, an approach based on notions of saliency has been recently investigated  in~\cite{Gonella_Haupt_IEEE_2013}.

The concept of sparsity is also at the core of recent efforts in \emph{dictionary learning} methods (e.g.,~\cite{Olshausen97, Lewicki98learningovercomplete, Kreutz03, Aharon06, Mairal_2010_OLM_1756006_1756008}), whose objective is to seek a sparse decomposition of a data set in terms of a basis (typically, but not necessarily, overcomplete) that is learned directly from the data. The idea of representing a signal using a small set of atoms of a learned dictionary instead of prescribed basis functions, such as sinusoids or wavelets,  has shown to be a powerful and versatile tool in many signal and image processing applications, including image denoising and restoration, texture synthesis and classification~\cite{mairal2008sparse, peyre2009sparse, ramirez2010classification}, audio classification~\cite{Zubair2013960} and source separation~\cite{Rambhatla:13}, and medical imaging~\cite{6469983}.

In this paper, we introduce a sparse coding approach to the structural diagnostics task based on the reinterpretation of the dictionary learning problem as a generalized, anomaly-sensitive form of modal analysis. We show that we can steer the outcome of the sparse coding problem toward the identification of anomalous features in the medium through the introduction of a data model whose parameters embody features associated with certain morphological, structural and/or behavioral characteristics that we presume should be exhibited in the response data. The resulting inference problem is fully model-agnostic and baseline-free, in that it does not require any \emph{a priori} knowledge of the material model of the medium (e.g., governing equations, material properties); therefore our proposed approach is well suited to study media whose material model is unknown, due for example to heterogeneity in the property distribution, or unreliable, as in the presence of material degradation extended over large portions of the domain. The approach is also database-free, as the construction of the data model is unsupervised, i.e., does not involve any training sets.

The remainder of the paper is organized as follows. In Section II we present the dictionary learning paradigm as a natural (and versatile) generalization of conventional modal analysis. In Section III we show how the sparse coding problem can be modified to capture localized features in the response data. The effectiveness of the method is demonstrated in Section IV using synthetic data as well as data acquired using a scanning laser Doppler vibrometer. In Section V we provide some final remarks and we discuss a few possible directions for future work.

\section{Dictionary Learning as Generalized Modal Analysis}
\begin{figure*}[t]
\centering
		\includegraphics[scale=0.65]{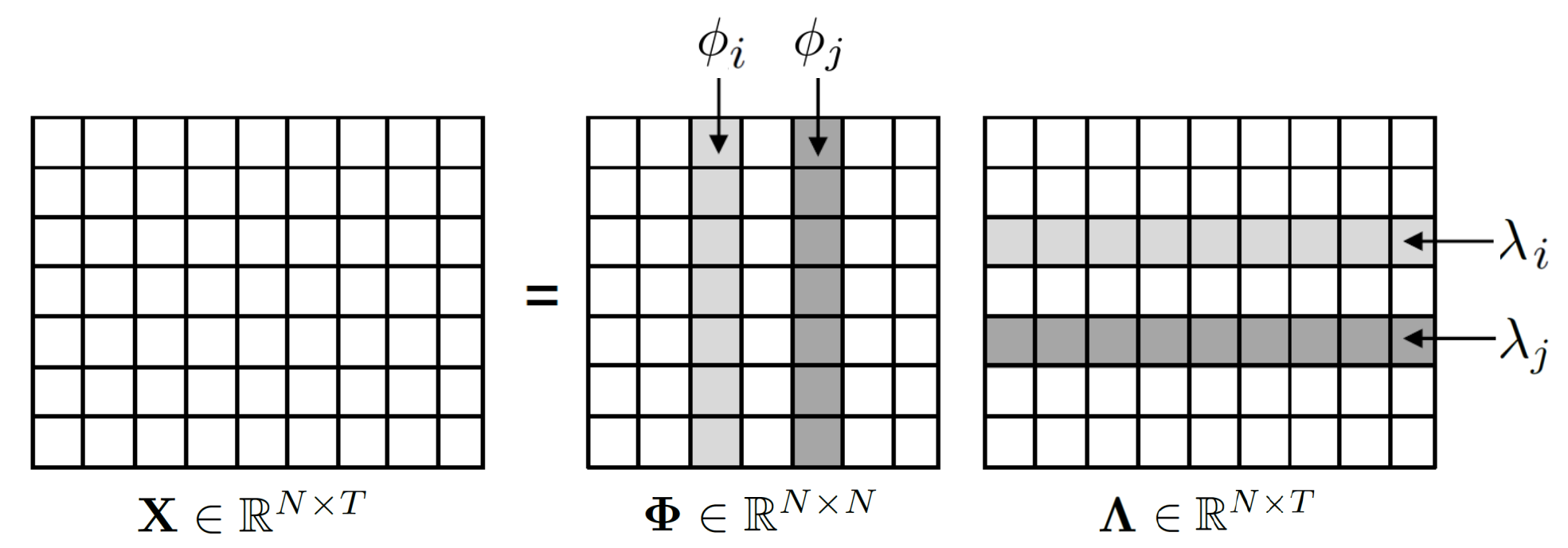}
	\caption{Matrix representation of conventional modal decomposition. Shaded columns of $\bold{\Phi}$ represent spatial functions associated with the digitized time functions appearing as shaded rows of $\bold{\Lambda}$. }
\label{Mat_Mult}
\end{figure*}

\indent In the context of structural dynamics, it is well known that the arbitrary motion of an undamped continuous system can be described as a linear combination of its natural modes, i.e., the dynamic displacement field $w(\bm{x},t)$ can be expressed as
\begin{equation}
\label{one}
w(\bm{x},t) = \sum_{j=1}^{\infty} \gamma_j \phi_{j}(\bm{x})e^{i \omega_j t}
\end{equation}
where $\phi_j(\bm{x})$ are the mode shapes, which are functions of the position vector $\bm{x}$, $\omega_j$ are the natural frequencies of the system, and $\gamma_j$ are modal amplitudes that are determined from the initial conditions of the problem.  If the summation in  Eq.~\ref{one} is truncated to $M$ terms, it yields an approximation of the displacement field. If we consider a discrete system with $N$ degrees of freedom (e.g., a lumped-parameter system or a continuous system upon spatial discretization), whose response is sampled at $T$ discrete time instants, we can update Eq.~\ref{one} and write the motion at a time instant $t_k$ as
\begin{equation}
\label{two}
\bold{w}(t_k) = \sum_{j=1}^{N} \gamma_j \bm{\phi}_{j} e^{i \omega_j t_k} \; \;  k=1,...,T. 
\end{equation}
where $\bold{w}$ is an $N \times 1$ array of degrees of freedom and $\bm{\phi}_j$'s are $N \times 1$ arrays representing the vectorized discrete mode shapes. We note that the exact response aggregates $N$ modal contributions, i.e., the number of modes equals the number of degrees of freedom. The modal decomposition is often recast in the form of a matrix multiplication as $\bold{X} = \bold{\Phi \Lambda}$, where $\bold{X} \in \mathbb{R}^{N \times T} $ is the matrix of the response data, containing a length $T$ discrete time history for each of the $N$ degrees of freedom, $\bold{\Phi} \in \mathbb{R}^{N \times N}$ is the (square) modal matrix whose $j^{th}$ column corresponds to the vectorized mode shape $\bm{\phi}_j$, $\bold{\Lambda} \in \mathbb{R}^{N \times T}$ is a matrix whose $\lambda_{j}$ row corresponds to the time-harmonic evolution of the $j^{th}$ modal coordinate, and the modal amplitudes $\gamma_j$ are  assumed absorbed into the rows of $\bold{\Lambda}$. In short, the modal decomposition can be expressed as a representation in terms of a purely spatial component (a matrix of discrete mode shapes) and a purely temporal one (a matrix of discrete harmonic functions). This decomposition is schematically illustrated in Fig.~\ref{Mat_Mult}.

As a generalization of this approach, we may wish to construct an alternative approximate representation of the response which is exclusively \emph{learned} from the response data (without the need for direct knowledge of the system's mechanical properties) but, unlike modal decomposition, relaxes \emph{a priori} assumptions on the mathematical form of the basis functions. In analogy with the description in terms of modes, we seek a representation that still involves a pair of spatial and temporal matrices, but in which the orthogonality constraint on the columns of the spatial matrix has been relaxed.  In compact form, we write
\begin{equation}
\label{approx_1}
 \bold{X \approx \bold{D}A},
 \end{equation}
where the matrix $\bold{D}$ is a \emph{dictionary} matrix, whose columns (called \emph{atoms}) form a set of spatial basis functions, and $\bold{A}$ is a corresponding coefficient matrix, whose rows may still be viewed as digitized functions of time. 
In this representation, the columns of $\bD$ play a role that is functionally analogous to the mode shapes $\phi_j$, therefore we refer to the atoms of $\bD$ as \emph{``pseudo-mode'' shapes} of the system.

Factorizations of the form of  Eq.\eqref{approx_1} are, of course, not unique without further qualifications; indeed, even the discretized form of standard modal decomposition may be written using this formalism. In what follows, we adopt this type of general factorization model in order to facilitate ``tuning'' of our learned data representations, so that they become inherently sensitive to wave propagation characteristics that are indicative of material defects.  We can encapsulate such characteristics in terms of the structure(s) that we \emph{prescribe to} or \emph{enforce upon} the learned factors $\bD$ and $\bA$.  Suppose that we specify classes of candidate dictionaries and candidate coefficient matrices, denoted by ${\cal D}$ and ${\cal A}$ respectively, so that each $\bD\in{\cal D}$ and each $\bA\in{\cal A}$ exhibits structural characteristics that we wish to impose on the dictionary atoms and their coefficients.  Then, given $\bX$, the aim of the representation task becomes to find specific factors $\widehat{\bD}\in{\cal D}$ and $\widehat{\bA}\in{\cal A}$, such that $\bX \approx \widehat{\bD} \widehat{\bA}$. Formally, our approach will be to identify these factors by solving a (constrained) version of a least-squares problem of the form
\begin{equation}\label{eqn:optDAgen}
\{\widehat{\bD}, \widehat{\bA}\} = \ \ \arg \min_{\bD\in{\cal D}, \bA \in {\cal A}} \ \ \sum_{t=1}^T\|\bx_t - \bD\ba_t\|_2^2,
\end{equation}
where $\bx_t$ and $\ba_t$ denote the $t$-th columns of $\bX$ and $\bA$, respectively, and the notation $\|\cdot\|_2^2$ denotes the squared $\ell_2$, or Euclidean, norm.

This line of thinking is motivated by recent efforts in \emph{dictionary learning} \cite{Olshausen97, Lewicki98learningovercomplete, Kreutz03, Aharon06, Mairal_2010_OLM_1756006_1756008} whose objectives are factorizations characterized by dictionaries that may be overcomplete (having more columns than rows) with corresponding coefficient matrices that are \emph{sparse} (having relatively few nonzero entries).  In terms of \eqref{eqn:optDAgen} above, dictionary learning tasks may be described as optimizations over a set $\cal{D}$ of matrices having $N$ rows and some user-specified number of columns (say $K$), and a corresponding set $\cal{A}$ of $K\times T$ coefficient matrices having no more than $s<K$ non zeros per column.  Enforcing sparsity on the columns of $\bA$ may be accomplished by imposing a set of constraints of the form $\|\ba_t\|_0 \leq s$ for all $t=1,2,\dots,T$, on the elements $\bA\in\cal{A}$, where the notation $\|\ba_t\|_0$ denotes the $\ell_0$ or counting norm of $\ba_t$, which essentially measures how many of its entries are nonzero\footnote{Strictly speaking, the function $\|\cdot\|_0$ does not satisfy all of the required characteristics for it to be a proper norm.  In particular, it fails the homogeneity property, in that for a vector $\ba$ we do not have $\|c\ba\|_0 = |c| \  \|\ba\|_0$ for all constants $c$. Nevertheless, it has become common in the sparse inference literature to use the norm descriptor for this function; we adopt the same convention here.}, and $s>0$ is a specified sparsity level.

Optimizations of this form (having $\ell_0$ constraints) are well-known to be combinatorial in nature. Thus, modern dictionary learning efforts either resort to greedy methods \cite{Aharon06}, or attempt to \emph{relax} the $\ell_0$ constraint on each column to yield computationally tractable constraints.  Often, this entails replacing the $\ell_0$ norm arising in the constraints on the columns of $\bA$ by its closest \emph{convex} surrogate~\cite{Mairal09}.  In terms of the terminology introduced for our problem, this latter approach would prescribe casting the dictionary learning task in terms of an optimization of the form
\begin{equation}\label{eqn:optDAl1}
\{\widehat{\bD}, \widehat{\bA}\} = \arg \min_{\bD\in{\cal D}, \bA \in {\cal A}} \ \ \sum_{t=1}^T \ \|\bx_t - \bD\ba_t\|_2^2 + \lambda \|\ba_t\|_1,
\end{equation}
where $\cal{A}$ is the set of all $K\times T$ matrices, $\cal{D}$ is the set of $N\times K$ matrices satisfying $\|\bd_j\|_2 \leq 1$ for each column $j=1,2,\dots, K$, and $\|\ba_t\|_1 \triangleq \sum_{i=1}^K |A_{i,t}|$ denotes the $\ell_1$ norm of $\ba_t$.  Here, $\lambda > 0$ is a (user-specified) regularization parameter that trades off a global ``goodness of fit'' of the approximation, as quantified by the $\ell_2$ term in the objective function, with the $\ell_1$ term on the columns of $\bA$; larger values of $\lambda$ tend to result in sparser $\bA$ having fewer non zeros per column. We now propose an extension of the (sparse) dictionary learning paradigm, in which we incorporate additional structural characteristics into the dictionary atoms to be identified, so that they be especially receptive to local deviations in the measured wavefields (and in turn, highly receptive to the wave propagation characteristics in the neighborhoods of local material anomalies).  We motivate and describe this approach in the following section.

\section{Dictionary Learning for Anomaly Detection}
\subsection{Dictionary learning for local feature extraction}

The objective of anomaly detection is to discover regions that behave abnormally, i.e., whose behavior significantly deviates from that of their surroundings. In the realm of structural diagnostics, the term \emph{anomaly} encompasses a wide variety of geometrical and material abnormalities, including damage zones, manufacturing defects and inclusions. In this discussion, we pursue the inference and triangulation of structural anomalies through the analysis of the structure's dynamic response, which consists of a collection of displacement time histories acquired at a set of discrete points on the structure's surface. Our contention is that it is possible to detect the signature of these anomalies by learning appropriate dictionaries of the response data and by decoding the spatial information contained in their data structure.

\begin{figure*}[t]
\centering
		\includegraphics[scale=0.65]{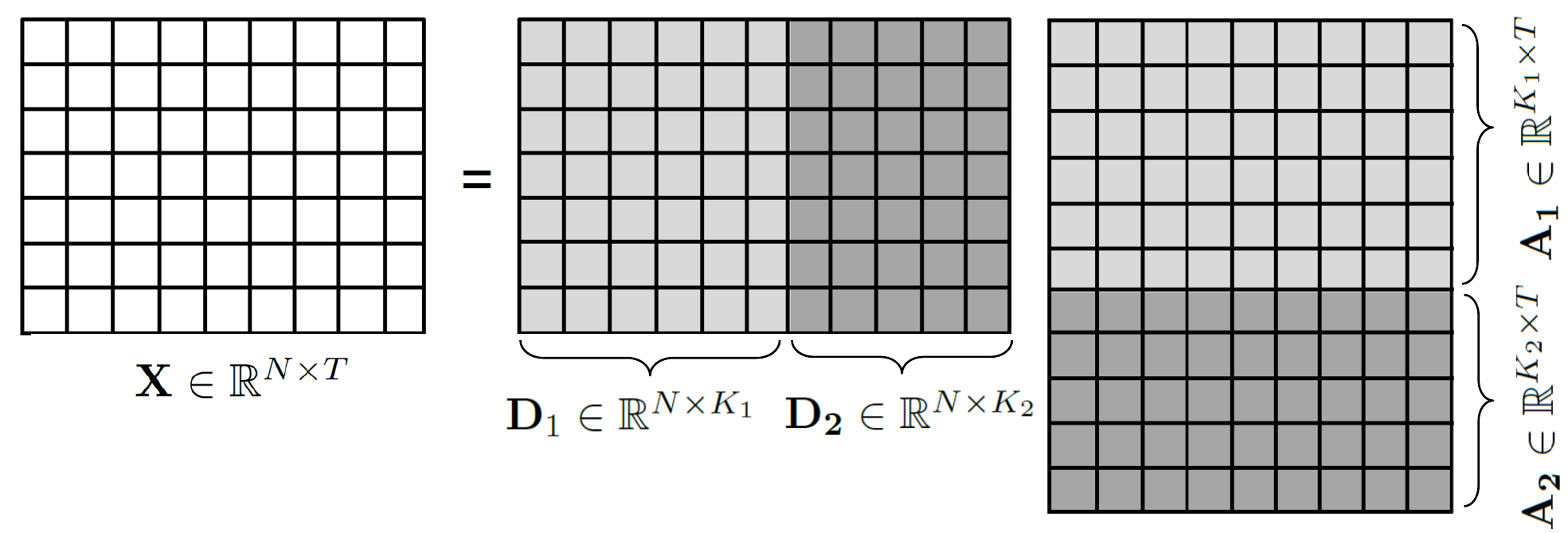}
	\caption{Schematic illustration of the two-dictionary approximate representation: $\bold{X} \approx \bold{D}_1\bold{A}_1 + \bold{D}_2\bold{A}_2$ with $\bold{D}_1\in \mathbb{R}^{N \times K_1}, \bold{D_2} \in \mathbb{R}^{N \times K_2}, \bold{A_1} \in \mathbb{R}^{K_1 \times T}$, $\bold{A_2} \in \mathbb{R}^{K_2 \times T}$.} 
\label{Mult_Mat}
\end{figure*}

We assume that the regions containing physical anomalies are characterized by local perturbations of their acoustic properties (e.g. elastic moduli, density); these, in turn, induce localized features in the response, which are reflected, although possibly difficult to detect, in the kinematic time histories of the material points which lie inside the anomalous regions. If we invoke the interpretation of the atoms of a dictionary as pseudo modes (i.e., \emph{spatial} descriptors of deformation), we can assume that the effect of localization would manifest as spiky regions in one or more of the atoms, which, from a data standpoint, would correspond to sparse structures in the columns of the dictionary. We conclude that, in order to equip a dictionary with the ability to detect anomalies, we need to formally enforce some kind of sparsity constraint on its atoms. On the other hand, the localized features associated with the anomalies coexist in the response with smoother fields describing the global (and dominant) behavior of the structure; therefore, a dictionary that properly captures an anomaly is unlikely to provide a sufficiently accurate representation of the response field as a whole, and viceversa. In order to reconcile this dichotomy, we propose a two-dictionary representation of the form
\begin{equation}
\label{approx}
 \bold{X \approx \bold{D}_1 A_1 + \bold{D}_2 A_2},
 \end{equation}
as illustrated in Fig.~\ref{Mult_Mat}.
This can be alternatively interpreted as a decomposition of $\bD$ into two separate dictionaries, so that $\bD = [\bD_1 \, \bD_2]$, where the $N\times K_1$ dictionary $\bD_1$ is dedicated to capture the localized features of the response and the $N\times K_2$ dictionary $\bD_2$ guarantees that the bulk response is sufficiently well approximated. Each dictionary is obtained through a dedicated optimization problem with appropriate constraints: specifically, for a user-specified constant $\Gamma>0$, we impose on the atoms of $\bD_1$ the sparsity-promoting constraint $\|(\bd_1)_j\|_2^2 + \Gamma\|(\bd_1)_j\|_1 \le 1$ for each column $j=1,2,\dots,K_1$, while we retain the original column-wise constraints from the original dictionary learning formulation on the columns of $\bD_2$ (that $\|(\bd_2)_j\|_2^2 \le 1$ for all $j=1,\dots,K_2$).  As we will see, the constraint on $\bold{D}_2$ tends to yield diffuse atoms, which \emph{de facto} display spatial smoothness, while the additional $\ell_1$-based constraint promotes sparsity on the columns of $\bD_1$. Together, $\bold{D}_1$ and $\bold{D}_2$ form a representation that encompasses the dominant, smooth dynamic behavior as well as the spatially sparse signature of potential anomalies. 
The search for $\bold{D_1, D_2, A_1, A_2}$ is done iteratively as detailed in Algorithm~\ref{algo:interations}.

\begin{algorithm}[t]
 \SetAlgoLined

 \DontPrintSemicolon 
\KwIn{Data cube $\bold{X} \in \mathbb{R}^{N \times T}$  }
\hspace{3.5em}Number of atoms in dictionaries: $K_1, K_2 > 0$\\
\hspace{3.5em}Regularization parameters: $\lambda, \Gamma >0$\\
\hspace{3.5em}Sparsity parameter $\varepsilon >0$\\
\hspace{3.5em}Increment parameter $\delta >0$\\
\KwOut{Dictionaries $\bold{D}_1$ (sparse) and $\bold{D}_2$ (diffuse)\vspace{2em}}

\textbf{Initialize:}
\begin{equation}
\begin{aligned}
& \underset{\bold{D_1} \in \mathbb{R}^{N \times K_1},\bold{A_{1}} \in \mathbb{R}^{K_1 \times T}} {\text{minimize}}
& & \sum_{t=1}^{T}||\bx_t -\bold{D}_1 (\ba_1)_t ||_2^2 + \lambda ||(\ba_1)_t||_1 \\
&\text{subject to}
& & \|(\bd_1)_j\|_2^2 + \Gamma\|(\bd_1)_j\|_1 \le 1 \quad \forall \quad 1\le j \le K_1 
\end{aligned}
\end{equation}

\While{$\|\bold{D}_1\|_0 \ge$  $\varepsilon$} {

\

$\Gamma = \Gamma + \delta$\\
$\bold{X}' = \bold{X} - \bold{D}_1\bold{A}_1$\\

\begin{equation}
\begin{aligned}
& \underset{\bold{D_2} \in \mathbb{R}^{N \times K_2},\bold{A_{2}} \in \mathbb{R}^{K_2 \times T}} {\text{minimize}}
& & \sum_{t=1}^{T}||\bx'_t -\bold{D}_2 (\ba_2)_t ||_2^2 + \lambda ||(\ba_2)_t||_1 \\
&\text{subject to}
& & \|(\bd_2)_j\|_2^2 \le 1 \quad \forall \quad 1\le j \le K_2
\end{aligned}
\end{equation}

$\tilde{\bold{X}} = \bold{X} - \bold{D}_2\bold{A}_2$\\

\begin{equation}
\begin{aligned}
\label{min_alg}
& \underset{\bold{D_1} \in \mathbb{R}^{N \times K_1},\bold{A_{1}} \in \mathbb{R}^{K_1 \times T}} {\text{minimize}}
& & \sum_{t=1}^{T}||\tilde{\bx}_t -\bold{D}_1 (\ba_1)_t ||_2^2 + \lambda ||(\ba_1)_t||_1 \\
&\text{subject to}
& & \|(\bd_1)_j\|_2^2 + \Gamma\|(\bd_1)_j\|_1 \le 1 \quad \forall \quad 1\le j \le K_1
\end{aligned}
\end{equation}


}
\Return{$\bold{D}_1,\bold{D}_2$}\;

\vspace{4mm}
 \caption{Iterative procedure to learn two dictionaries (\emph{sparse} and \emph{diffuse}) from the response data.}\label{algo:interations}
\end{algorithm}

In the minimization problem of Eq.~\ref{min_alg}, the constraint on the individual columns of dictionary $\bD_1$ are of the form $||(\bd_1)_j||_2^2 + \Gamma||(\bd_1)_j||_1 \le 1$. Note that the $\Gamma$ parameter effectively governs the number of non-zero terms in each atom. The larger $\Gamma$, the more stringent the sparsity constraint and the more zero elements in each atom, which in return makes the spatially sparse features more prevalent in the atoms of $\bold{D}_1$. However, if $\Gamma$ is too large, the constraint becomes excessively stringent and the cost function is trivially minimized with $||\bold{D}_1||_0 = 0$. On the other hand, if $\Gamma$ is too small, the constraint  is not sufficiently invoked and the atoms of $\bold{D}_1$ effectively lose the ability to highlight any localized features. The determination of the appropriate value of $\Gamma$ is computed iteratively starting with a low initial guess $\Gamma_0$ and increasing it by an amount $\delta$ until the columns of $\bold{D_1}$ have a sufficiently low number $\varepsilon$ of non-zero entries. In our implementation, the solution of the individual dictionary learning problems involved at each stage of iteration is carried out using the open-source sparse modeling software SPAMS (available at \verb"http://spams-devel.gforge.inria.fr").

Let us recall at this point that the role of the regularization parameter $\lambda$ in the cost function of each dictionary learning problem is to control the trade-off between the accuracy of the approximation and the achievement of a parsimonious dictionary representation involving few atoms. In the present implementation, the selection of $\lambda$ has been conducted following a heuristic approach such that the error in the approximation remained below acceptable bounds ($\approx 10 \%$ of $\bold{X}$).

\subsection{A super-atom approach to capture persistence and enhance anomaly detection}

Important parameters of the algorithm are the user-defined parameters $ K_1$ and $K_2$, which represent the assumed number of atoms in $\bold{D}_1$ and $\bold{D}_2$, respectively. If we let $K_1$ be large, we have the opportunity to examine a high number of sparse atoms. This richness of sparse descriptors can be exploited to better determine if the system contains an anomaly. It is, in fact,  possible that some sparse atoms may not display the ``true" physical anomalies, but rather other spurious short-wavelength features, e.g. cusps associated with boundary effects, which may manifest as localized features on the perimeter of  the structure, or sharp artifacts due to noise. Nevertheless, we note that the true anomalies are a \emph{persistent} feature in the sparse dictionary, i.e., features that are consistently observed across the set of sparse atoms and at consistent locations within each atom. In order to capture this \emph{persistence} attribute, we propose a post-processing aggregation step designed to intelligently aggregate layers of data from multiple atoms in a way that emphasizes the most persistent features. The result of this step is a kind of \emph{super-atom} that highlights spatial locations where persistent activity is present across a significant number of the sparse dictionary atoms. The construction of the super-atom proceeds as follows. We consider a partition of the domain into $M_1 \times M_2$ rectangular (identical in size and shape) regions. Since the atoms spatially span the entire domain, they are all partitioned in similar fashion, such that $\bold{d}_j^{k}$ denotes the $k^{th}$ partition of the $\bold{d}_j$ atom. For each partition, we sweep the atoms of the sparse dictionary and we check if a feature is consistently observed in that partition across the set, by counting how many atoms contain at least one non zero entry inside the selected partition. If this number is sufficiently large, we aggregate local contributions from all the atoms in the dictionary to form the corresponding partition of the super-atom. Note that this criterion weighs (possibly relatively mild) contributions that are observed over a large number of atoms more heavily than others that may be prominent (amplitude wise), but are observed only in a few atoms. This reflects the notion that  the signature of physical anomalies is often elusive but persistent across the dictionary, while spurious sharp features, which can dominate the response amplitude wise, are inconsistently detected across the dictionary. The construction of the super-atom (summarized in Algorithm~\ref{algo:superatom}) features two parts: the first implements the atom aggregation procedure; the second performs a search over the identified partitions, with the objective of automatically identifying, through amplitude thresholding, the partition containing the anomalies. This last step is meant to forsake the need for visual inspection of the super-atom and \emph{de facto} makes the anomaly identification fully automatic.

\begin{algorithm}[t]
 \SetAlgoLined

\DontPrintSemicolon 
\KwIn{Sparse Dictionary $\bold{D}_1=[\bold{d}_1 \ \bold{d}_2 \ \dots \ \bold{d}_{K_1}] \in \mathbb{R}^{N \times K_1}$  }
\hspace{3.5em}Domain partition sizes: $M_1, M_2 > 0$\\
\hspace{3.5em}Sparsity thresholds $\varepsilon, \delta >0$\\
\vspace{2mm}
\KwOut{``Super-atom'' $\bold{S}$ (with vectorized partitions $\{\bold{S}^k\}$)}
\vspace{2mm}
\textbf{Initialize:} \ $M=M_1 M_2$; $Q=N/M$\\
\hspace{5.5em}$\bold{S}^k= \bold{0} \in \mathbb{R}^Q$ for $k=1,\dots,M$\\
\vspace{2mm}
\textbf{Partition:} Divide each $\bold{d}_j$, $j=1,\dots,K_1$, into length $Q$ sub vectors $\{\bold{d}_j^k\}_{k=1}^M$

\vspace{2mm}

\For{$k = 1$ \textbf{to} $M$} {
 \If{$ \displaystyle\sum\limits_{j=1}^{K_1} \mathbf{1}_{\{||\bold{d}_j^k||_0>0\}} \ge$  $\varepsilon$}
   { $\bold{S}^k \gets \displaystyle\sum\limits_{j=1}^{K_1} ||\bold{d}_j^k||_0$}

\For{$ i = 1$ \textbf{to} $Q$}{
    \If{$|S^k_i| \le$  $\delta$}
   { $S^k_i \gets 0 $}
}
}

\Return{$\{\bold{S}^k\}$}\;
\vspace{4mm}
 \caption{Construction of super-atom aggregating contributions from different sets of sparse atoms according to criteria of persistence. The function $\mathbf{1}_{\{\cdot\}}$ denotes the \emph{indicator function}, which takes the value $1$ when the event specified in the subscript is true, and $0$ otherwise.  
 }\label{algo:superatom}
\end{algorithm}

Note that there may be instances where the algorithm does not find any partition that contains any sufficiently persistent localized features; this scenario corresponds to the case of a pristine structure. The ability to avoid false positives in anomaly-free scenarios is a major strength of this methodology, in contrast with methods that do not explicitly control for false positives, e.g. our own previous work~\cite{Gonella_Haupt_IEEE_2013}, (which utilized a  simpler, principal components analysis-based anomaly detection and  localization method).

The benefits of a super-atom representation are fully realized when we consider problems in which the identification of an anomaly is impossible, or ambiguous, even through the prism of the sparse atom dictionary. This scenario is encountered when the signature of the anomaly is very small, due, for example, to a critically small size of the defect or to competition (in terms of sparsity) from spurious response features, or in dealing with noisy data. In short, the signature of an anomaly may be minute, and hence not visually detectable, in the individual atoms, but can become very clear through its aggregation over many atoms.

It is worthwhile to note that our super-atom  post-processing method is a bit of a departure from existing dictionary learning approaches, which typically utilize directly the atoms identified by the learning procedure(s) without further refinement. Our motivation for adopting this additional step is twofold. First, as described above, the approach seeks to identify spatially persistent features in the response data by aggregating (in a nonlinear manner) features identified in the atoms of the sparse dictionary. In addition, we note that while dictionary learning problems are easy to motivate and specify, their highly non-convex nature  makes their exact numerical solution computationally challenging. In practice, existing algorithmic approaches rely either on convex relaxation and  alternating optimization or greedy methods and can only be guaranteed to converge to local minima of the corresponding objective function. In this sense, our post processing step may be viewed as an augmentation designed to glean additional information from the computational solution of the dictionary learning problem.

\section{Local Feature Extraction in a Transient Wavefield}

In this section we provide several numerical demonstrations of the proposed anomaly identification method, on both simulated data and experimental data obtained using a scanning laser Doppler vibrometer.

\subsection{Results from numerical simulations}

First we test the approach against the problem of circular-crested transversal waves excited in a thin plate by an out-of-plane point force applied at one node. We consider flexural waves modeled according to Mindlin's plate theory. The choice of flexural waves is here primarily motivated by the simplicity of the corresponding finite element simulation and the inherent simplifications resulting from having a single-mode wave solution. Nevertheless, since the method is based on the elaboration of the spatial patterns of time-evolving wavefields, without invoking any specific physical characteristics of the waves, the analysis would hold for other types of out-of-plane waves, such as Lamb waves and Rayleigh waves, and even for longitudinal and shear waves in thin structures exhibiting some in-plane deformation on the structure's surface. We use the finite element method (FEM) to construct the stiffness and mass matrices of the system and employ a time marching scheme to simulate the propagating wavefield. Our virtual specimen is a rectangular, thin Aluminum sheet with dimensions $L_x= 1.0 \textrm{m}, L_y=0.5 \, \textrm{m}$ and thickness $h=0.002 \, \textrm{m}$ and the following material parameters: Young's modulus  E = 71GPa, Poisson ratio $\nu$ = 0.33, density $\rho = 2700 \, \textrm{Kg}/\textrm{m}^\textrm{3}$. The domain is discretized with a structured mesh comprising 400 $\times$ 200 square elements, which is verified to be sufficiently fine to avoid spurious numerics-induced dispersive effects or artificial noise in the data. The excitation frequency is a 5-cycle tone burst with carrier frequency $f_c=100\textrm{KHz}$, which induces a wavefield with wavelength $\lambda$ such that $L_x/ \lambda \approx 30$. At the end of simulation, our data cube $\bold{X}$ consists of a series of snapshots of the propagating wavefield at the selected $T$ time instants.

The anomalies are introduced in the model by relaxing the Young's modulus of the material by two orders of magnitude within small regions of the domain. This can model the effects of partial holes, soft inclusions or localized regions with degraded material properties. The anomalies behave as scatterers, which act as localized sources within the domain that are triggered after some delay with respect to the applied excitation. As the point source of excitation is itself a localized feature, we expect it to pose some ambiguity for the sparse coding algorithm. In order to (partially) filter out this effect and discriminate between the ``true" anomalies and the excitation, we truncate the first $25\%$ time instants of the response (corresponding to the early stages of propagation), in which we expect the response to be dominated by the excitation as the wavefield is localized in the neighborhood of the excitation point. For the same reason, a $0.01L_x $-thick layer immediately close to the boundary where the excitation is applied, is \emph{a priori} excluded from the analysis.

 \begin{figure}[h!]
\centering
   \begin{center}
    \subfloat[Diffuse atoms ($9$ random samples of $16$-atom dictionary)]{\label{single_diffuse}\includegraphics[scale=0.6]{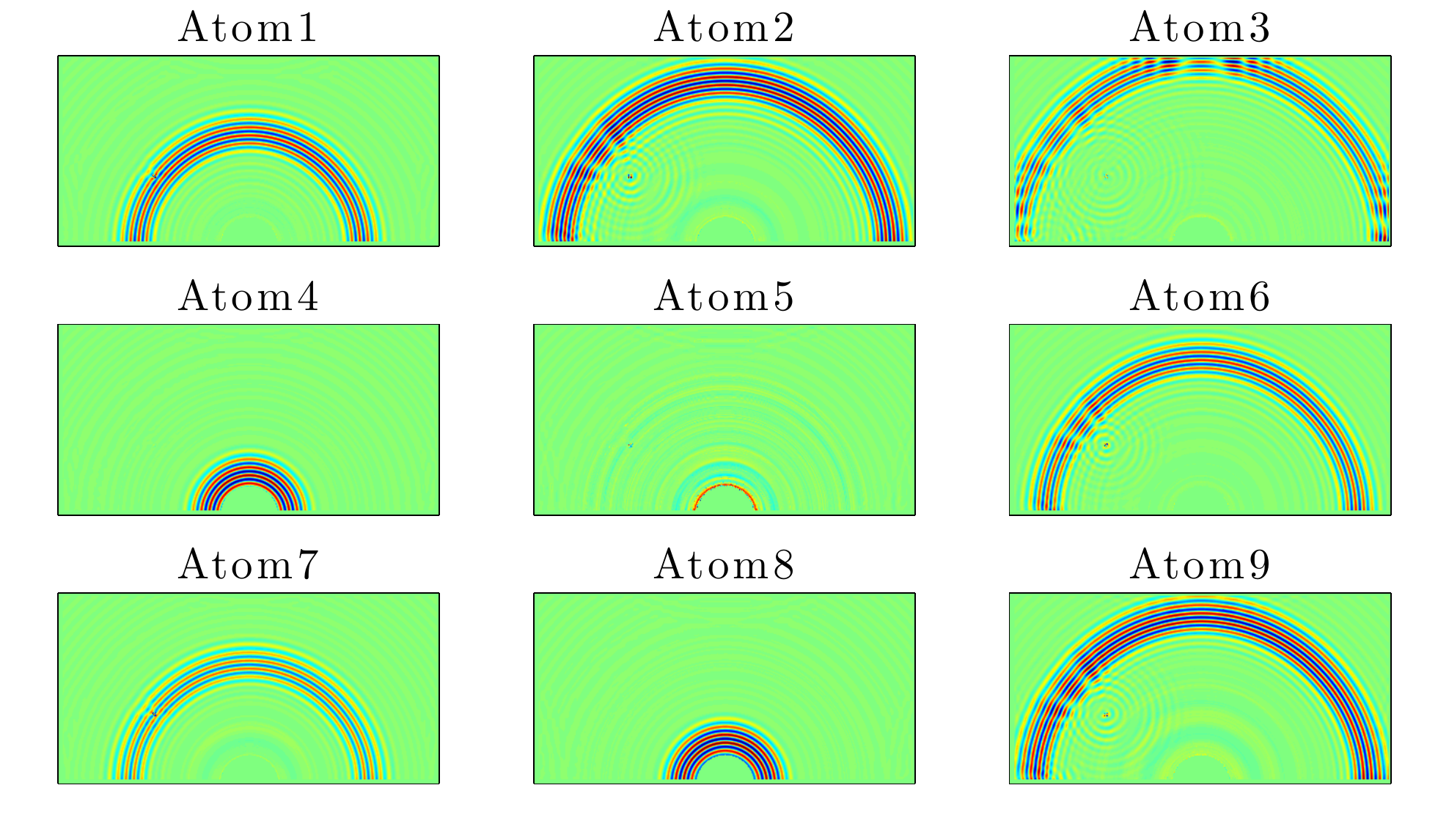}} \hfill   
    \subfloat[Sparse atoms ($9$ random samples of $100$-atom dictionary)]{\label{single_sparse}\includegraphics[scale=0.6]{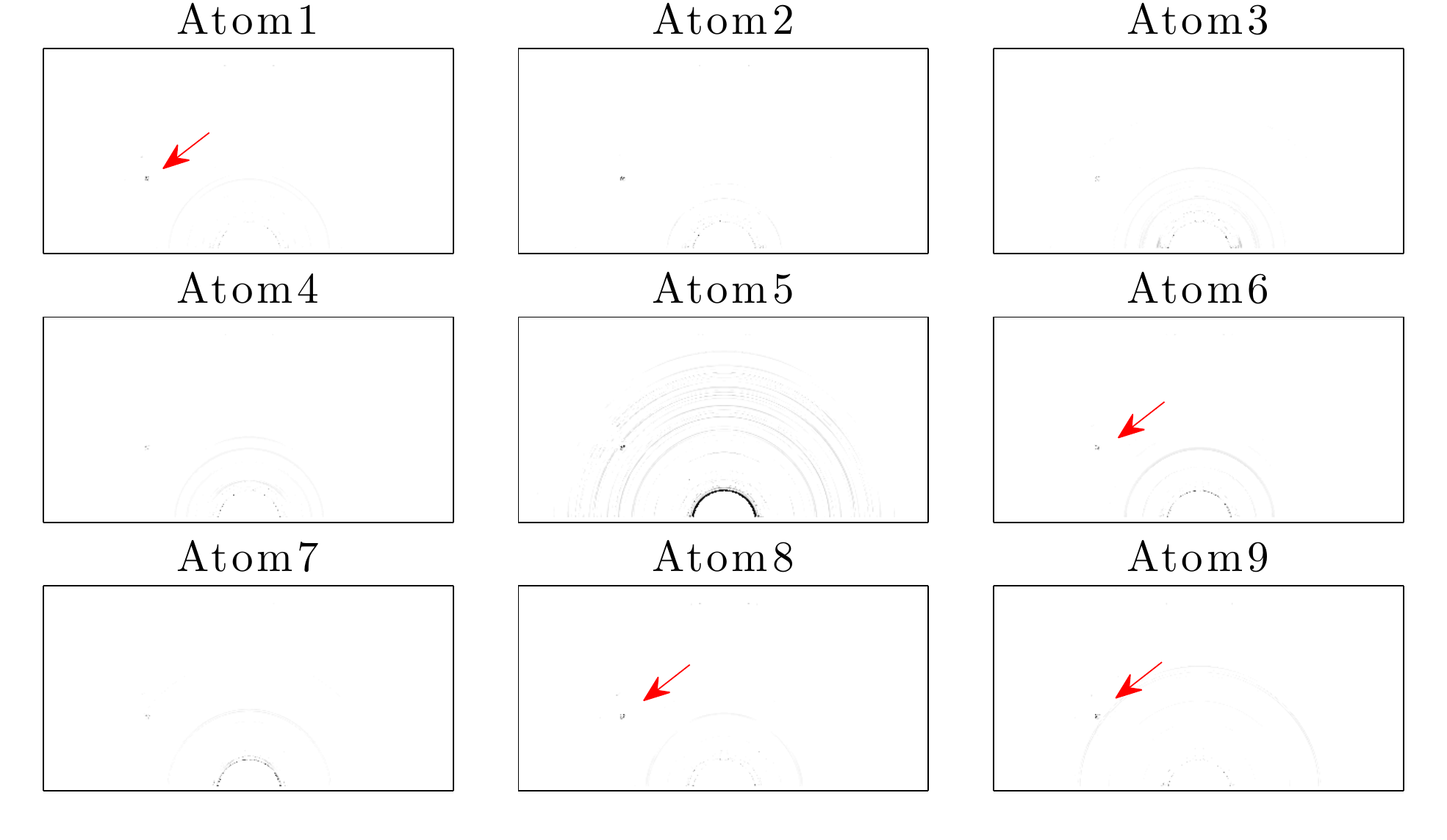}} \\  
   \end{center}
       \caption{Samples of atoms from diffuse and sparse dictionaries for a numerically generated wavefield with point anomaly. Arrows are used to assist the visualization of the spikes corresponding to the anomaly.}
  \label{single_dictionaries}
\end{figure}

\begin{figure}[h!]
\centering
   \begin{center}
    \subfloat[Schematic of plate with point defect]{\label{single_schematic}\includegraphics[scale=0.6]{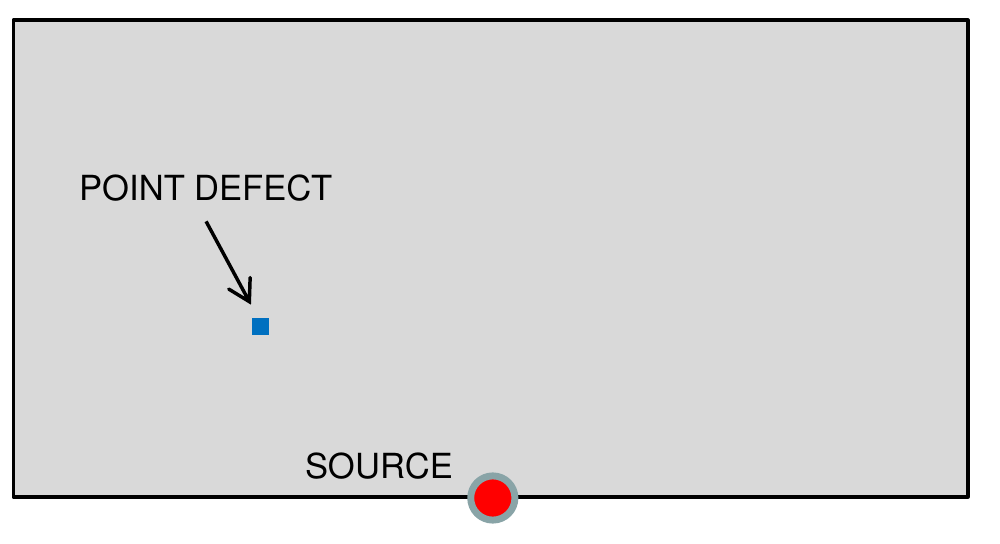}} 
   \hspace{0.2in}
    \subfloat[Super-atom highlighting anomalous region]{\label{single_superatom}\includegraphics[scale=0.6]{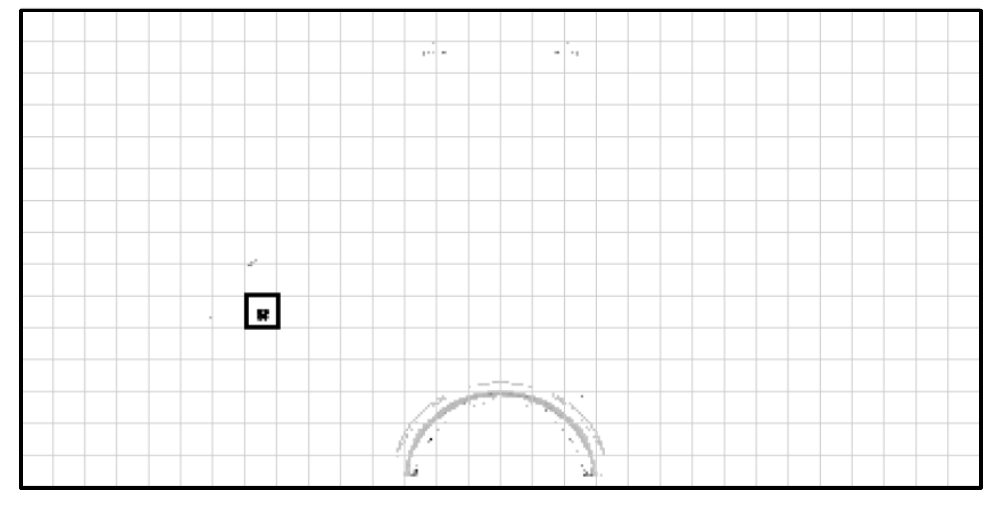}} \\
    \subfloat[Schematic of plate with two point defects]{\label{double_schematic}\includegraphics[scale=0.6]{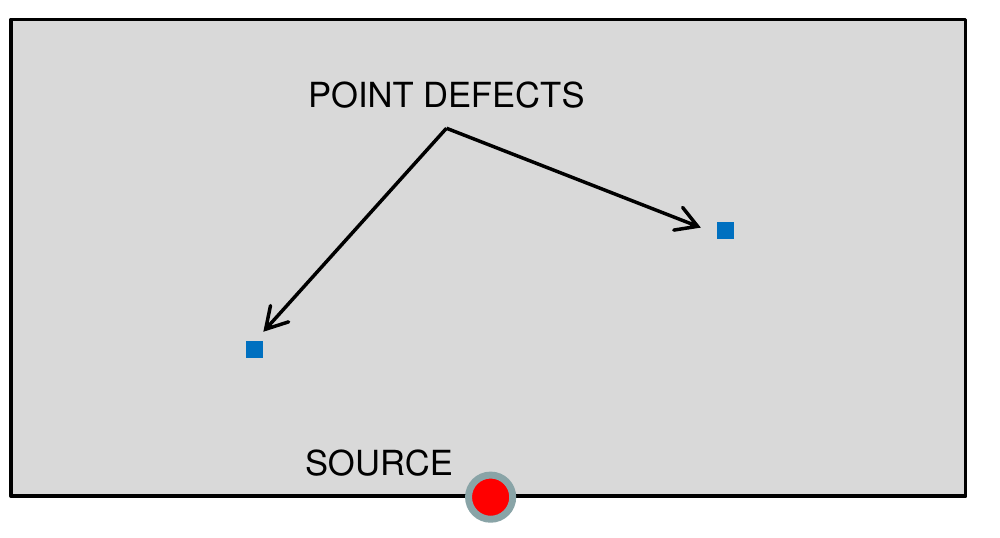}} 
   \hspace{0.2in}
    \subfloat[Super-atom highlighting anomalous regions]{\label{double_superatom}\includegraphics[scale=0.6]{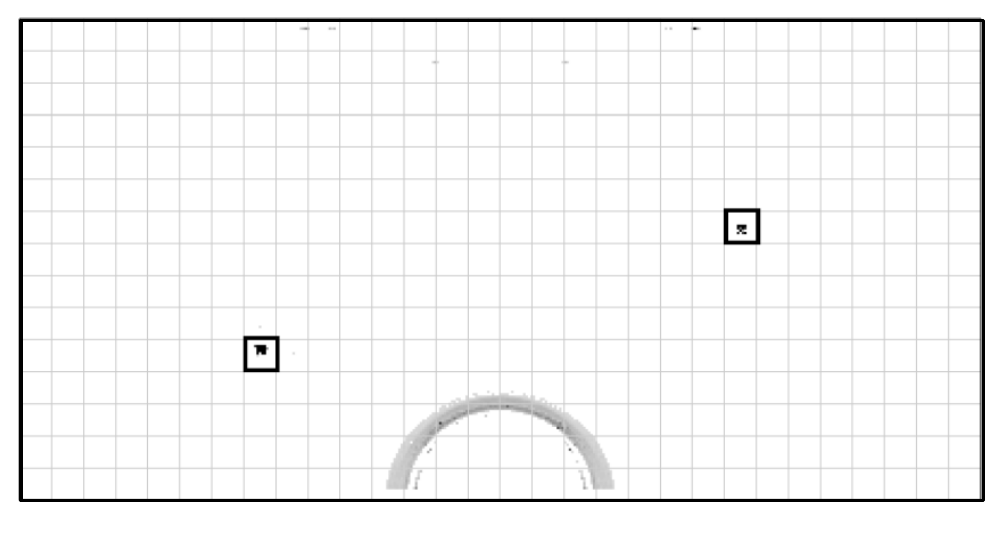}} \\
        \subfloat[Schematic of plate with point and line defect]{\label{crack_schematic}\includegraphics[scale=0.6]{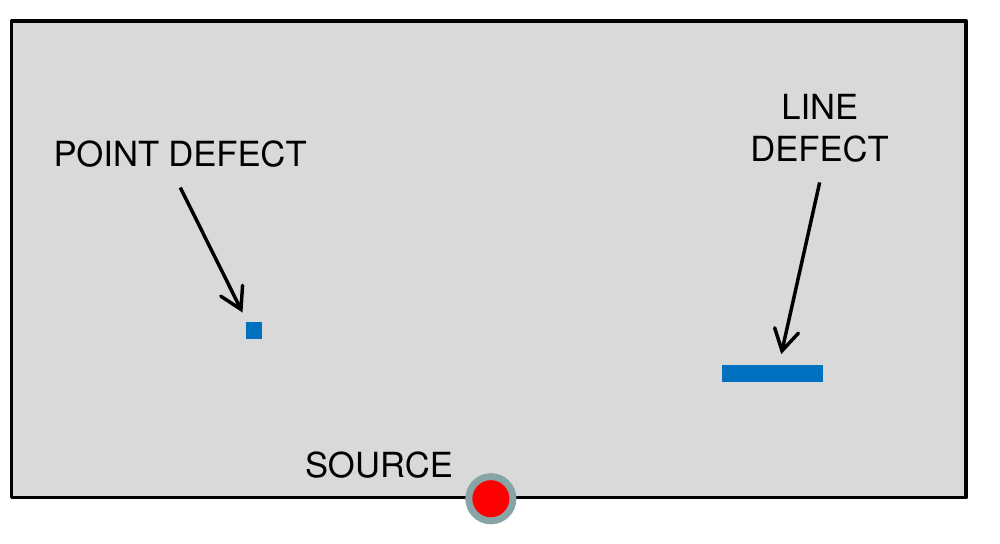}} 
   \hspace{0.2in}
    \subfloat[Super-atom highlighting anomalous regions]{\label{crack_superatom}\includegraphics[scale=0.6]{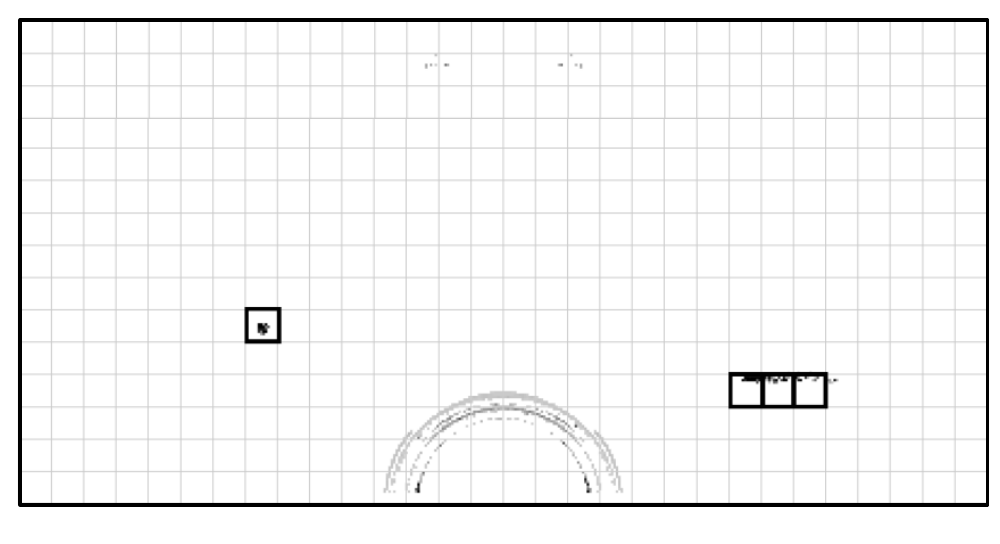}} \\
   \end{center}
       \caption{Anomaly detection and triangulation through the prism of super-atom representation. Point and line anomalies are successfully extracted from the wavefields and triangulated.}
  \label{first_detection}
\end{figure}

In Fig.~\ref{single_dictionaries} we show a sample of 9 atoms of a low-dimensional ($K_2=16$) diffuse dictionary $\bold{D}_2$ and a sample of 9 randomly selected atoms of a higher-dimensional ($K_1=100$) sparse dictionary $\bold{D}_1$ for a case with a single point anomaly of coordinates $(0.25 L_x, \, 0.33 L_y)$. The diffuse atoms (Fig.~\ref{single_diffuse}) essentially capture characteristic snapshots of the wavefield at different instants of simulation. In contrast, the sparse dictionary (Fig.~\ref{single_sparse}) captures a spatially rarefied representation of the wavefield, in which several atoms display a distinguishable and isolated signature of the anomaly in the form of a localized spike in their topology; this signature is, however, quite elusive, as the anomalous feature is not ubiquitously observed across the entire set and the task of discriminating it from other speckles in each atom is prohibitive.  As expected, the inference benefits vastly from a super-atom representation (Fig.~\ref{single_superatom}), in which the sparse features are weighted according to their persistence across the entire sparse dictionary. The sporadically occurring features that contaminate the sparse atoms are now filtered out and we are able to clearly pinpoint the anomaly, as visible from the comparison with the schematic of Fig.~\ref{single_schematic}. We note that the anomaly can even be triangulated without visual inspection as its host partition is automatically identified by the algorithm (and highlighted in the figure by a thicker border). In the remaining portion of Fig.~\ref{first_detection} we further explore the performance of the method against cases with more challenging anomaly landscapes.  In Fig.s~\ref{double_schematic} and~\ref{double_superatom} we show the super-atom performance for a plate with two defects, where both anomalies are correctly detected and triangulated. The final case in Fig.s~\ref{crack_schematic} and~\ref{crack_superatom} tests whether a crack, which in this case can be thought of as a contiguous collection of anomalies, can be identified even in the presence of another scatterer. The crack is simulated by reducing the Young's modulus and density of the material by several orders of magnitude inside a two-element thick horizontal layer extending between points $(0.74 L_x, \, 0.20 L_y)$ and $(0.84 L_x, \, 0.20 L_y)$. The signature of the crack is well highlighted in the super-atom; if we relax the search criterion and we let the algorithm find the 4 most anomalous partitions, we are able to automatically identify the entire length of the defect.

\subsection{Results from experimental data}

We now test the efficacy of the dictionary learning algorithm against experimental data through the benchmark problem of a thin plate with a localized defect, the physical analogue of the numerical case presented in the previous subsection. Two cases, using a similar specimen but with considerably different types of anomalies, are presented. Here we consider a square Aluminum sheet of dimensions $L_x=L_y=0.61 \textrm{m}$ and thickness $h=0.0394 \textrm{m}$ and we apply a tone burst excitation with carrier frequency $f_c=200 \textrm{kHz}$. The induced wavefield is reconstructed from surface velocity data via a Scanning Laser Doppler Vibrometer (SLDV), the Polytec PSV-400-3D. Three scanning heads are used to shoot three independent laser beams onto each node of a predefined scanning grid, which allows capturing the out-of-plane and in-plane velocity components of the points on the plate's surface; in this work, we limit our analysis to the out-of-plane velocity component. 
The plate surface is discretized with an approximately square scanning grid consisting of 120 $\times$ 120 scan points. The grid is deliberately chosen to be very fine to enable resolution of minute features in the wavefield; experiments with coarser grids can be easily performed \emph{a posteriori} by subsampling the data in space. To make the wavefield data more amenable to region partitioning, we apply linear interpolation to fit the original grid data to a Cartesian grid. Since surface roughness can cause signal dropouts, or temporary severe signal attenuation, built-in signal enhancement and speckle tracking features are enabled during the scan.  Moreover we average 75 realizations at each scan point to filter optical and/or mechanical noise (signal stacking). Despite many precautions taken during the scan, the acquired signals are tainted by some mechanical and optical noise; the data is therefore cleaned by post-processing the time history signals with a band pass filter with a 200 kHz center frequency and a bandwidth of 100 kHz. Time spacing between acquisitions at different points is enforced by a repetition frequency of $9 \textrm{Hz}$, which allows for over 1000 boundary reflections between successive acquisitions and effectively allows for full attenuation of the wave before a new acquisition is made.

\begin{figure}[h!]
\centering
   \begin{center}
    \subfloat[Diffuse atoms ($9$ random samples of $16$-atom dictionary)]{\label{hole_diffuse}\includegraphics[scale=0.36] {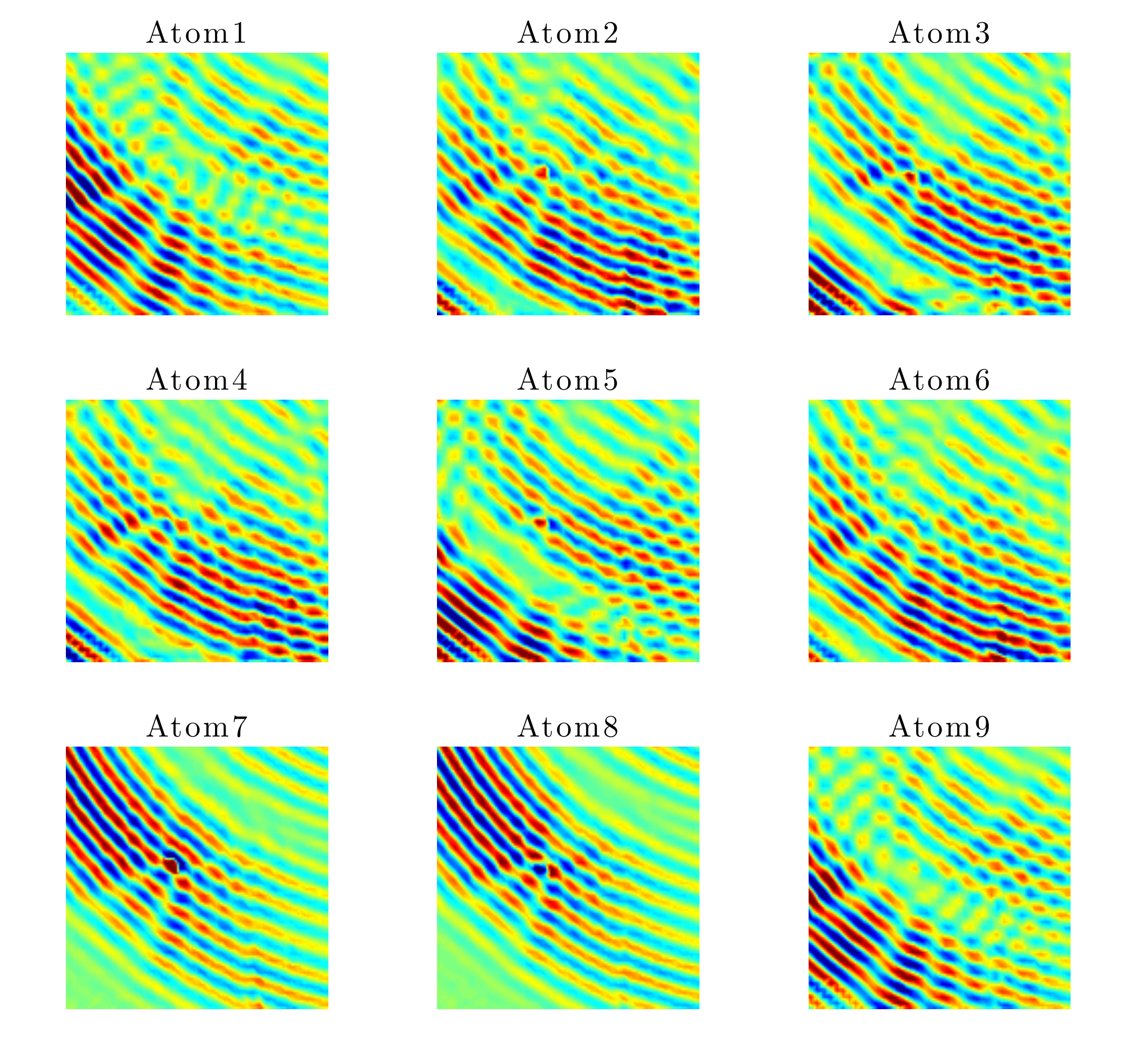}} \hfill
    \subfloat[Sparse atoms ($9$ random samples of $100$-atom dictionary)]{\label{hole_sparse}\includegraphics[scale=0.36] {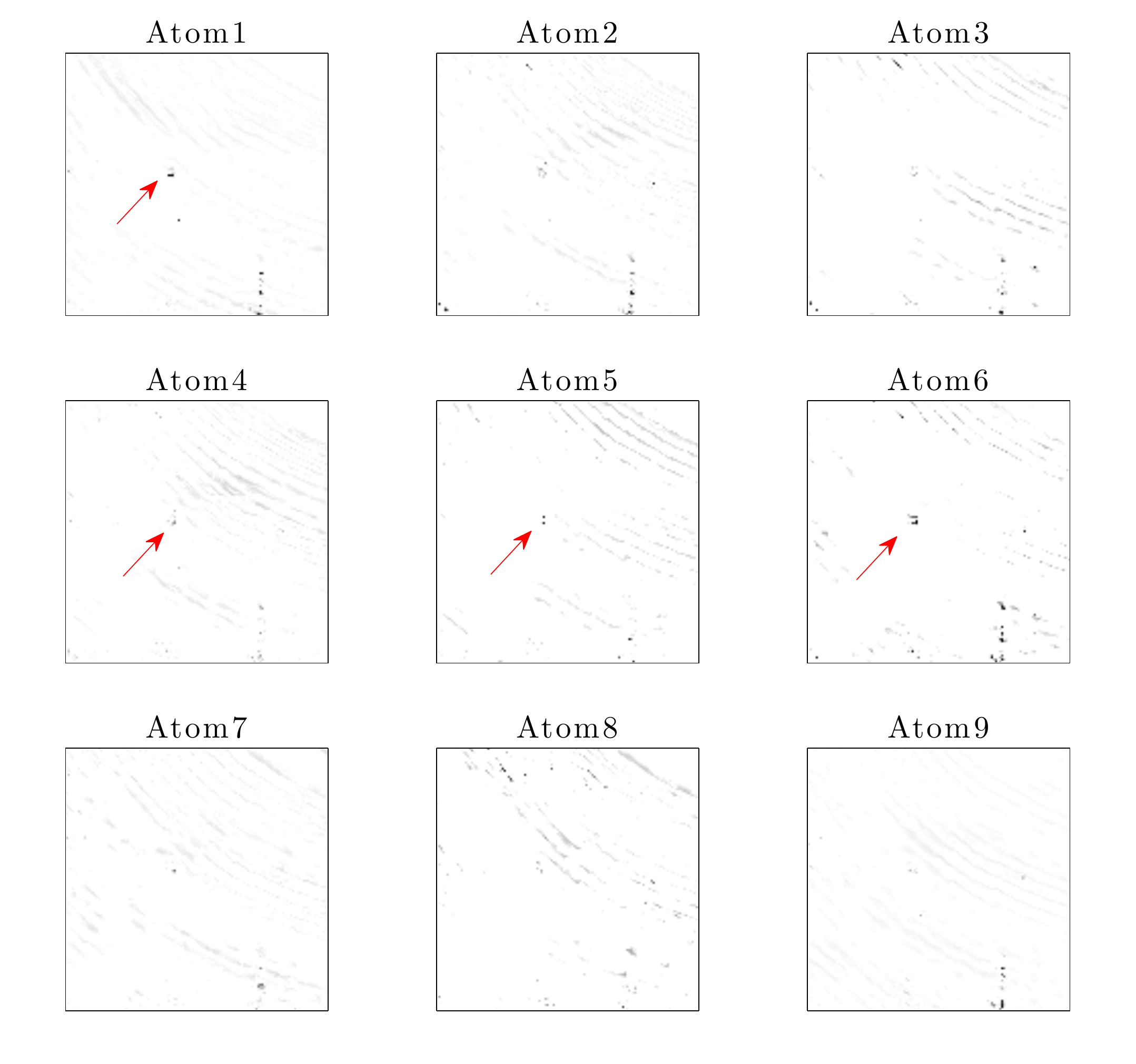}} \\
   \end{center}
       \caption{Samples of atoms from diffuse and sparse dictionaries for a laser-acquired wavefield with strong scatterer. Arrows are used to assist the visualization of the spikes corresponding to the anomaly.}
  \label{hole_dictionaries}
\end{figure}

\begin{figure}[h!]
\centering
   \begin{center}
    \subfloat[Schematic of plate with highlighted anomaly (picture detail) and scanned area]{\label{hole_schematic}\includegraphics[scale=0.75]{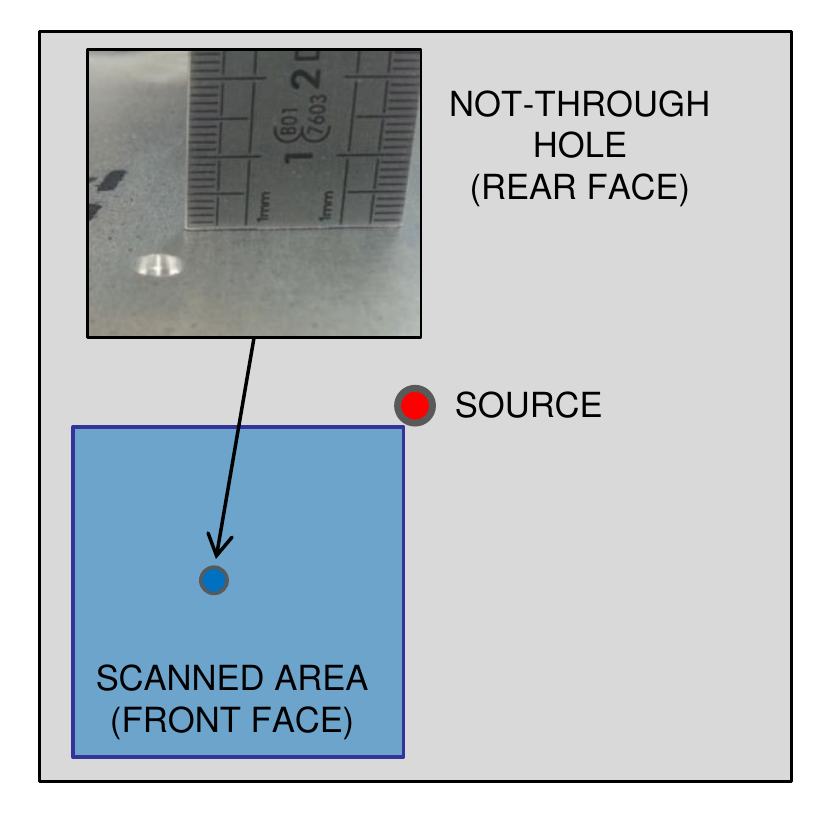}} 
   \hspace{1in}
    \subfloat[Super-atom of the scanned area successfully identifying the anomaly]{\label{hole_superatom}\includegraphics[scale=0.75]{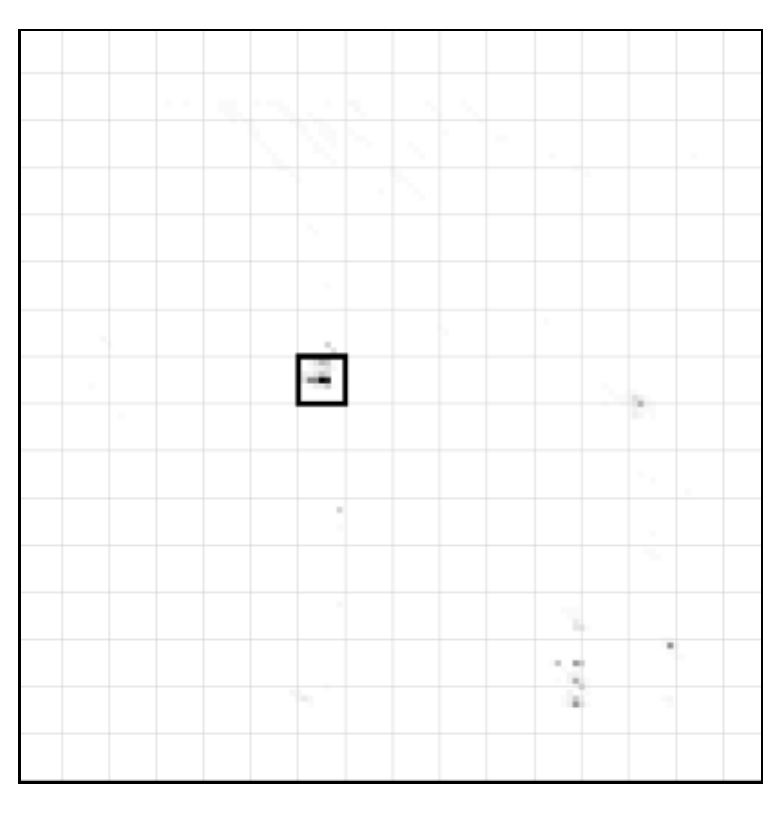}} \\
   \end{center}
       \caption{Anomaly detection and triangulation through the prism of super-atom representation. The defect corresponding to the hole is successfully triangulated.}
  \label{hole_detection}
\end{figure}

In the first experimental case we consider a localized anomaly introduced in the form of a circular, flat-bottomed cylindrical not-through hole, shown in detail in the schematic of Fig.~\ref{hole_schematic}. The hole is drilled on the back surface of the plate to a depth of approximately 0.0295 m - corresponding to a 75$\% $ reduction in local thickness - and is expected to behave as a strong scatterer. Nevertheless, it is worth pointing out that the scan is performed over a portion of the \emph{pristine} front face of the plate (also highlighted in Fig.~\ref{hole_schematic}), which shows no visual evidence of the defect; the experiment mimics diagnostics conditions encountered in the inspection of certain thin-walled structures - such as mounted aircraft panels, pipes, etc. - for which it may be prohibitive to have direct optical access to the interior surfaces. In Fig.~\ref{hole_dictionaries} we show the diffuse atoms and a randomly selected sub-sample of 100 computed sparse atoms. Similar to the numerical case, the diffuse atoms (Fig.~\ref{hole_diffuse}) effectively display snapshots of the wavefield at different time instants of the experiment, while the sparse atoms (Fig.~\ref{hole_sparse}) distill the sparse, or localized features of the response. While the presence of a distinguishable localized feature is observed in several of the atoms at the location of the anomaly, the inference is again contaminated by other sparse features, which introduces an element of ambiguity. Through the lens of the super-atom, as seen in  Fig.~\ref{hole_superatom}, the ``true" anomaly is clearly displayed and highlighted.

\begin{figure}[h!]
\centering
   \begin{center}
    \subfloat[Diffuse atoms ($9$ random samples of $16$-atom dictionary)]{\label{glue_diffuse}\includegraphics[scale=0.37]{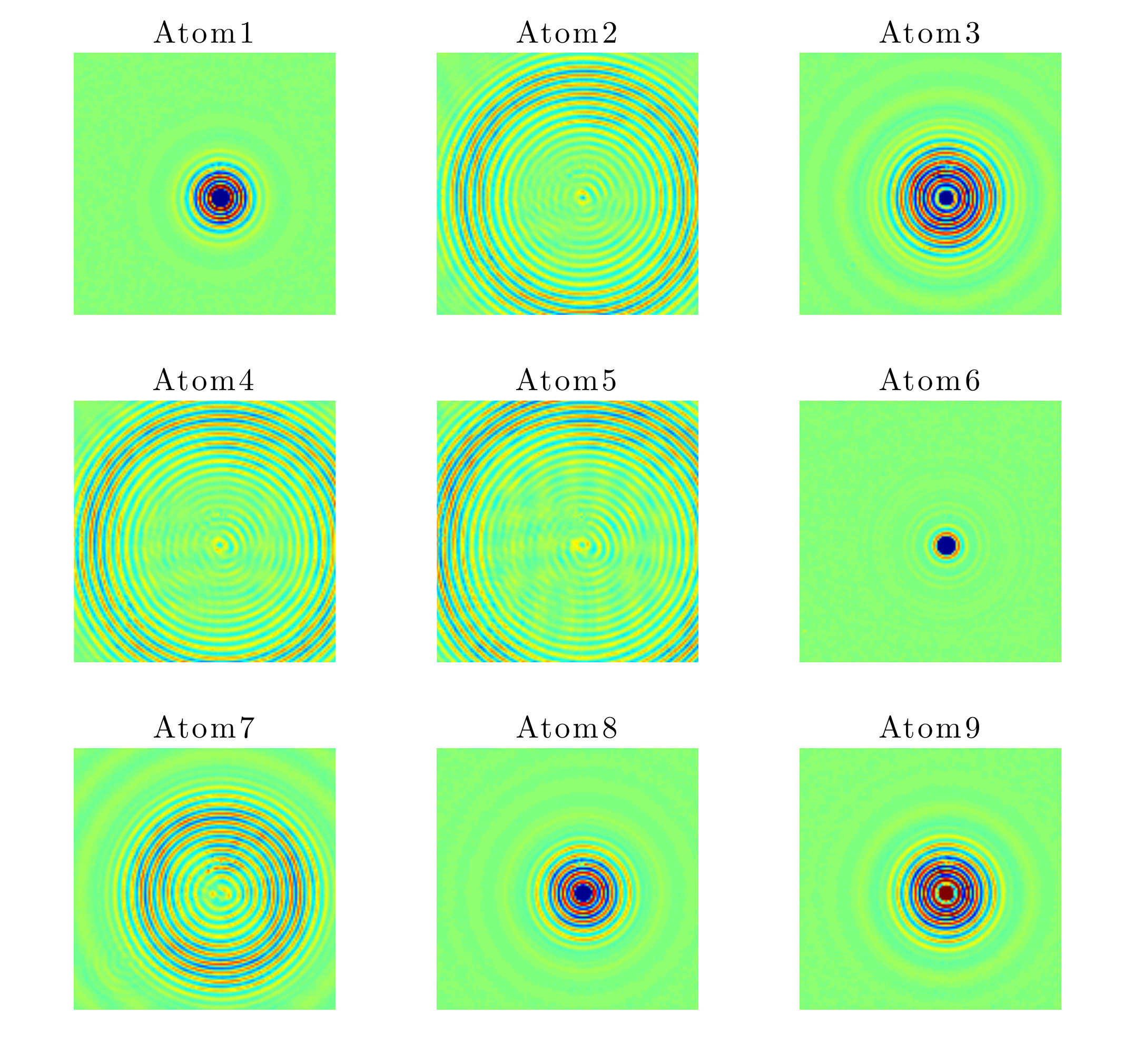}} \hfill
    \subfloat[Sparse atoms ($9$ random samples of $100$-atom dictionary)]{\label{glue_sparse}\includegraphics[scale=0.37]{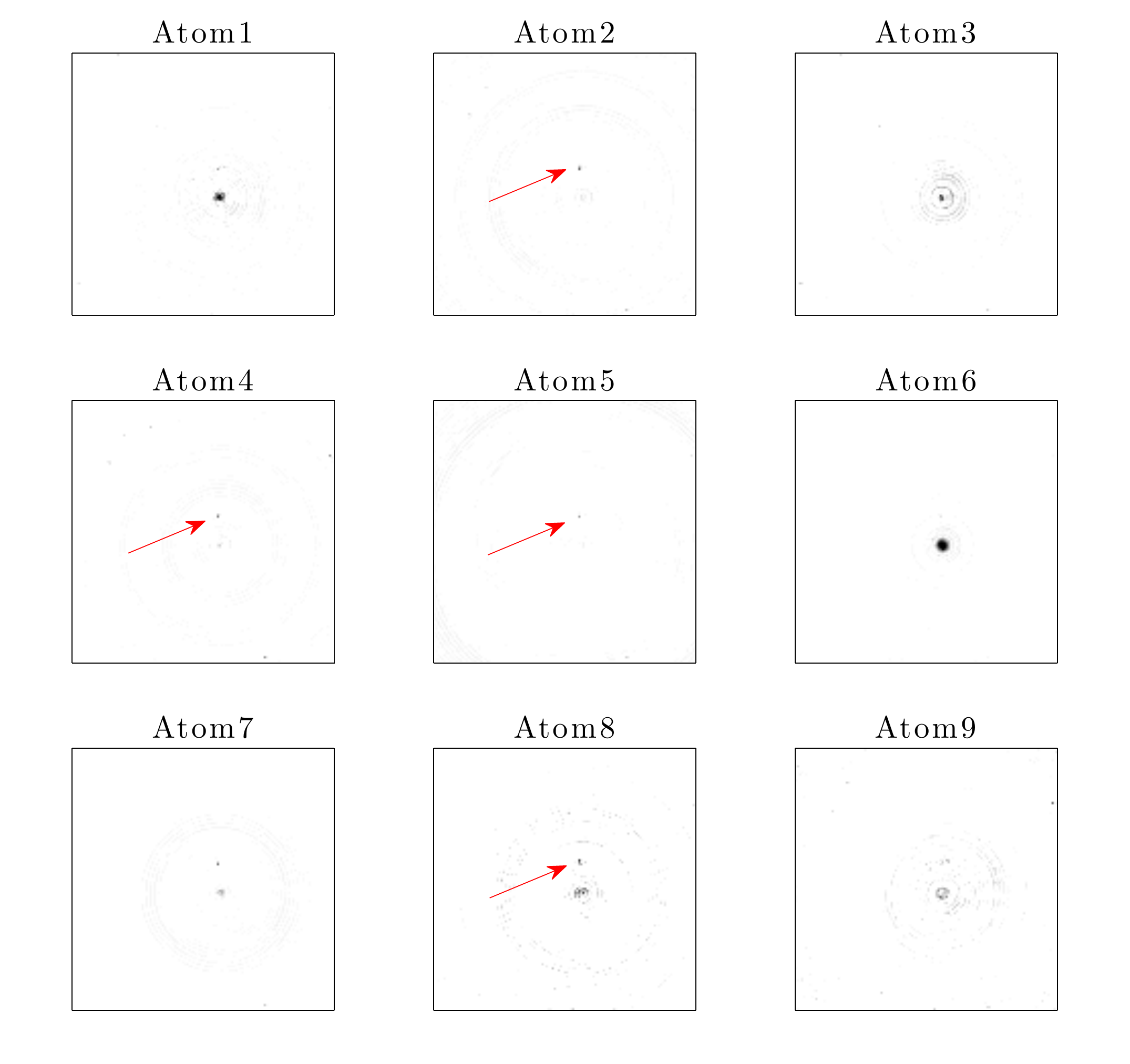}} \\
   \end{center}
       \caption{Samples of atoms from diffuse and sparse dictionaries for a laser-acquired wavefield with weak scatterer. Arrows are used to assist the visualization of the spikes corresponding to the anomaly.}
  \label{glue_dictionaries}
\end{figure}

\begin{figure}[h!]
\centering
   \begin{center}
    \subfloat[Schematic of plate with highlighted anomaly and scanned area]{\label{glue_schematic}\includegraphics[scale=0.75]{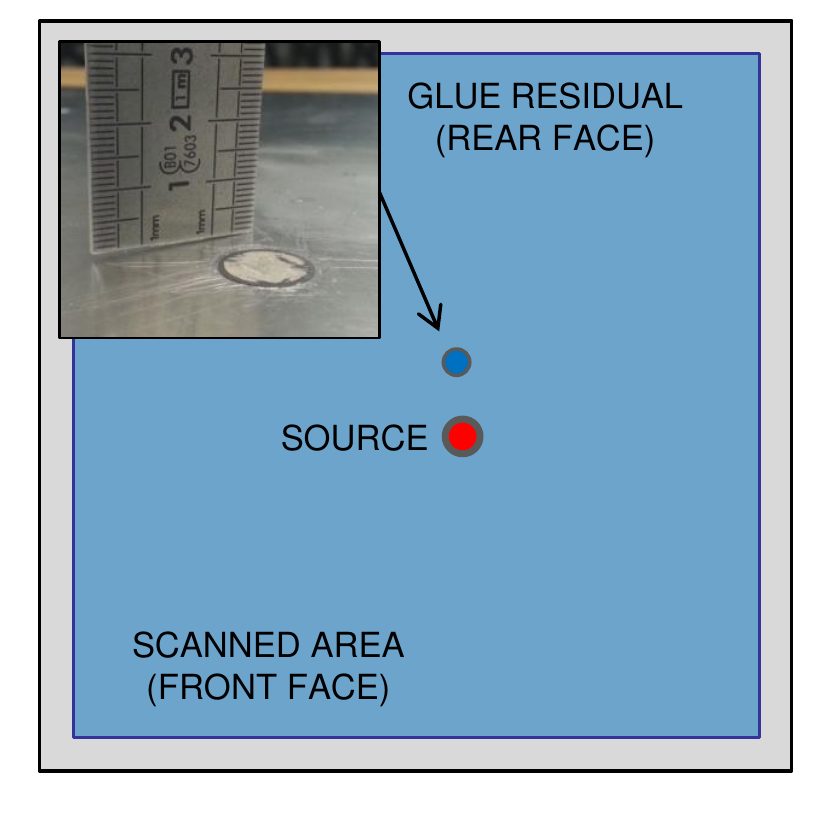}} 
   \hspace{1in}
    \subfloat[Superatom of the scanned area identifying source and anomaly]{\label{glue_superatom}\includegraphics[scale=0.75]{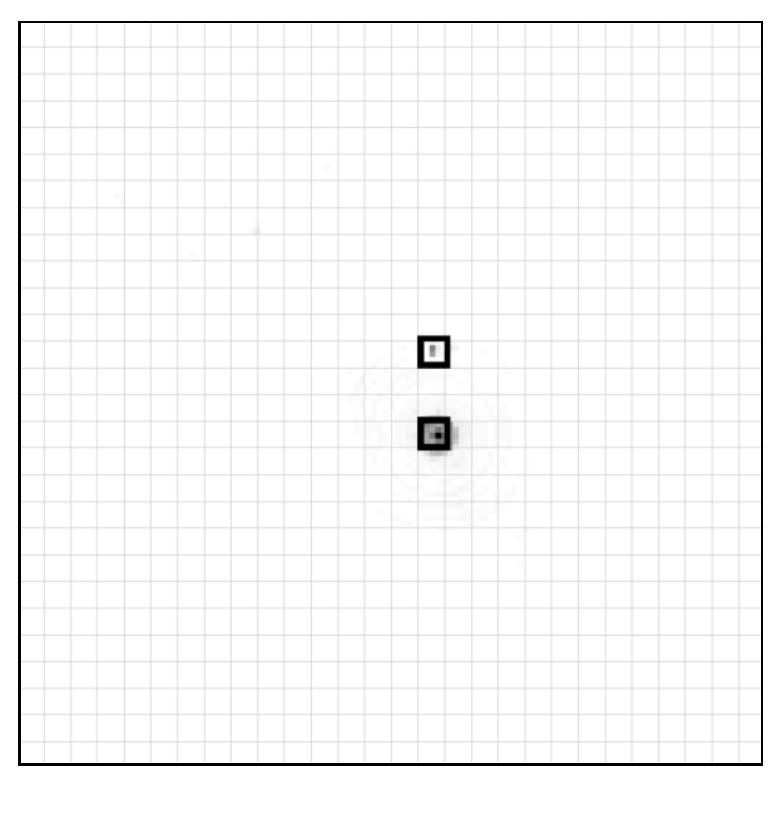}} \\
   \end{center}
       \caption{Anomaly detection and triangulation through the prism of super-atom representation. The defect corresponding to the glue residual is successfully triangulated.}
  \label{glue_detection}
\end{figure}

Our second experimental case is an intriguing testimony to the method's ability to detect subtle changes induced in a wavefield by minute or superficial defects. The almost serendipitous way in which this detection was obtained speaks volumes about the agnostic capabilities of the method. A plate with identical specifications as above (without the drilled hole), unknowingly possessed a very thin, approximately 0.1mm-thick deposit of glue - remnant of an actuator that had been previously glued to its back surface (see picture detail in the schematic of Fig.~\ref{glue_schematic}).  A preliminary scan was performed for the purpose of calibrating the equipment and obtaining a preliminary characterization of the wavefield. Since we were ostensibly dealing with a pristine plate, we expected a radially symmetric, unperturbed wavefield. Even though some marginal distortion potentially ascribable to a weak scatterer was detected upon visual inspection, the observed wavefield initially confirmed the prediction of an essentially pristine structure. However, when the data was fed to the sparse coding algorithm, a number of localized features were displayed in the sparse atoms (Fig.~\ref{glue_sparse}). One of them corresponds to the excitation point, which here lies in the middle of the domain - a region that is not \emph{a priori} excluded from the analysis; this is not surprising, as excitation sources are inherently spatially localized features. However, we also noted that many of the sparse atoms contained a second feature above the point of excitation. This was emphasized in the super-atom representation (Fig.~\ref{glue_superatom}), which indeed suggests the presence of a (weak) scatterer. Further inspection of the plate back surface revealed the presence of the aforementioned defect at the location indicated by the super-atom.

\section{Concluding Remarks and Future Work}

In this work we have introduced a methodology to detect and triangulate anomalies in the response of solid media based on criteria of spatial sparsity and enabled by sparse coding algorithms. The method represents, to the best of our knowledge, the first attempt to use structurally-tuned dictionary learning algorithms in the context of structural diagnostics and, in general, one of the few existing efforts aimed at performing anomaly detection from spatially reconstructed wavefields using techniques of computer vision and image processing. We have shown that the method can identify spatially localized anomalies in the data fields through a decomposition of the response in two sets of pseudo modes: diffuse ones, which capture the smooth part of the response, and sparse ones, which contain the signature of the anomalies. The sparse coding algorithms have been complemented with the assembly of super-atoms that intelligently aggregate the information from the sparse dictionaries to filter the localized features that are due to physical anomalies from those due to noise or other competing boundary-induced or numerics-induced mechanisms.

The method is crowned by a postprocessing feature that allows a convenient virtual decomposition of the domain in rectangular partitions for automatic identification of the regions containing the anomalies. The benefits of an automatic interpretation that bypasses the need for direct visual observations of the wavefield (or of the atoms of its dictionaries) are felt in the  context of possible \emph{multi-step} sampling and detection procedures, where the inference would be made in several stages conducted over nested sub-domains, and with increased accuracy, to iteratively identify smaller and smaller subsets of the material domain that may contain anomalies.  This sort of adaptive ``coarse-to-fine'' sampling strategy would enable agile and fast sensing and detection procedures which could in return enhance the applicability and competitiveness of image--processing-based diagnostics methods.  This is the objective of current investigations; an account of this is left for future work.

Another interesting avenue for future work consists of testing the approach in the context of heterogeneous materials, in which the material properties could potentially feature large deviations from an ideal case even far from the to-be-detected anomalies. This type of scenario would ultimately testify to the agnostic properties of the approach, as in those cases the ability to triangulate anomalies without precognition of the mechanical properties of the medium would fully manifest. 


\bibliographystyle{unsrtnat}
\bibliography{NSF_bib}

\begin{thebibliography}{54}
\providecommand{\natexlab}[1]{#1}
\providecommand{\url}[1]{\texttt{#1}}
\expandafter\ifx\csname urlstyle\endcsname\relax
  \providecommand{\doi}[1]{doi: #1}\else
  \providecommand{\doi}{doi: \begingroup \urlstyle{rm}\Url}\fi

\bibitem[Staszewski et~al.(2004)Staszewski, Boller, and Tomlinson]{Staszewski}
W.J. Staszewski, C.~Boller, and G.~Tomlinson.
\newblock \emph{Health monitoring of aerospace structures: Smart sensors and
  signal processing}.
\newblock Wiley \& Sons, 2004.

\bibitem[Rose(2002)]{Rose}
J.L. Rose.
\newblock A baseline and vision of ultrasonic guided wave inspection potential.
\newblock \emph{Journal of Pressure Vessel Technology}, 124\penalty0
  (3):\penalty0 273--282, 2002.

\bibitem[Ihn and Chang(2008)]{Pitch_Catch_Fu_Kuo_Chang}
J-.B. Ihn and F-.K. Chang.
\newblock Pitch-catch active sensing methods in structural health monitoring
  for aircraft structures.
\newblock \emph{Structural Health Monitoring}, 7\penalty0 (1):\penalty0 5--19,
  2008.

\bibitem[Flynn et~al.(2011)Flynn, Todd, Wilcox, Drinkwater, and
  Croxford]{Likelihood_Guided_Waves_Flynn}
E.~B. Flynn, M.~D. Todd, P.~D. Wilcox, B.~W. Drinkwater, and A.~J. Croxford.
\newblock Maximum-likelihood estimation of damage location in guided-wave
  structural health monitoring.
\newblock \emph{Proceedings of the Royal Society A: Mathematical, Physical and
  Engineering Science}, 467\penalty0 (2133):\penalty0 2575--2596, 2011.

\bibitem[Michaels et~al.(2005)Michaels, Michaels, Mi, and
  Ruzzene]{Michaels_Sparse}
T.~E. Michaels, J.~E. Michaels, B.~Mi, and M.~Ruzzene.
\newblock Damage detection in plate structures using sparse ultrasonic
  transducer arrays and acoustic wavefield imaging.
\newblock In \emph{AIP Conference Proceedings}, volume 760, page 938, 2005.

\bibitem[Liu et~al.(2012)Liu, Liu, and Yuan]{SVD_Sensors_Damage_NCState}
L.~Liu, S.~Liu, and F.-G. Yuan.
\newblock Damage localization using a power-efficient distributed on-board
  signal processing algorithm in a wireless sensor network.
\newblock \emph{Smart Materials and Structures}, 21\penalty0 (2):\penalty0
  025005, 2012.

\bibitem[Lu et~al.(2008)Lu, Wang, Tang, and Ding]{SVD_Sensors_TAMU}
Y.~Lu, X.~Wang, J.~Tang, and Y.~Ding.
\newblock Damage detection using piezoelectric transducers and the {L}amb wave
  approach: {II}. {R}obust and quantitative decision making.
\newblock \emph{Smart Materials and Structures}, 17\penalty0 (2):\penalty0
  025034, 2008.

\bibitem[Wang and Yuan(2009)]{Wang_Yuan_PZT_Locations_Damage_Detection}
Q.~Wang and S.~Yuan.
\newblock Baseline-free imaging method based on new {PZT} sensor arrangements.
\newblock \emph{Journal of Intelligent Material Systems and Structures},
  20\penalty0 (14):\penalty0 1663--1673, 2009.

\bibitem[Prada and Fink(1994)]{Prada94}
C.~Prada and M.~Fink.
\newblock Eigenmodes of the time reversal operator: {A} solution to selective
  focusing in multiple-target media.
\newblock \emph{Wave Motion}, 20\penalty0 (2):\penalty0 151--163, 1994.

\bibitem[Foroozan and Asif(2011)]{Foroozan_localization_algorithms}
F.~Foroozan and A.~Asif.
\newblock Time reversal based active array source localization.
\newblock \emph{IEEE Transactions on Signal Processing}, 59\penalty0
  (6):\penalty0 2655--2668, 2011.

\bibitem[Kessler et~al.(2002)Kessler, Spearing, and
  Soutis]{Kessler_Soutis_Lamb_insitu}
S.S. Kessler, S.M. Spearing, and C.~Soutis.
\newblock Damage detection in composite materials using lamb wave methods.
\newblock \emph{Smart Materials and Structures}, 11\penalty0 (2):\penalty0 269,
  2002.

\bibitem[Kirikera et~al.(2011)Kirikera, Balogun, and
  Krishnaswamy]{Kirikera_Balogun_Krishnaswamy_SHM}
G.R. Kirikera, O.~Balogun, and S.~Krishnaswamy.
\newblock {Adaptive Fiber Bragg Grating Sensor Network for Structural Health
  Monitoring: Applications to Impact Monitoring}.
\newblock \emph{Structural Health Monitoring}, 10\penalty0 (1):\penalty0 5--16,
  2011.

\bibitem[Farrar and Worden(2013)]{Farrar_ML}
C.~Farrar and K.~Worden.
\newblock \emph{Structural health monitoring: {A} machine learning
  perspective}.
\newblock Wiley \& Sons, 2013.

\bibitem[Zhongqing and Ye(2004)]{Lamb_waves_delamination_Neural}
S.~Zhongqing and L.~Ye.
\newblock Lamb wave-based quantitative identification of delamination in
  {CF}/{EP} composite structures using artificial neural algorithm.
\newblock \emph{Composite Structures}, 66\penalty0 (1):\penalty0 627 -- 637,
  2004.

\bibitem[{Lu} and {Michaels}(2007)]{Michaels_Matching_Pursuit}
Y.~{Lu} and J.~E. {Michaels}.
\newblock {Ultrasonic Signal Decomposition via Matching Pursuit with an
  Adaptive and Interpolated Dictionary}.
\newblock In D.~O. {Thompson} and D.~E. {Chimenti}, editors, \emph{Review of
  Progress in Quantitative Nondestructive Evaluation}, volume 894 of
  \emph{American Institute of Physics Conference Series}, pages 579--586, 2007.

\bibitem[Das et~al.(2005)Das, Papandreou-Suppappola, Zhou, and
  Chattopadhyay]{Das_et_al_MPD_SPIE}
S.~Das, A.~Papandreou-Suppappola, X.~Zhou, and A.~Chattopadhyay.
\newblock On the use of the matching pursuit decomposition signal processing
  technique for structural health monitoring.
\newblock \emph{Smart Structures and Materials 2005: Smart Structures and
  Integrated Systems}, 5764\penalty0 (1):\penalty0 583--594, 2005.

\bibitem[Das et~al.(2009)Das, Kyriakides, Chattopadhyay, and
  Papandreou-Suppappola]{Das_Chattopadhyay_MoteCarlo_MPD}
S.~Das, I.~Kyriakides, A.~Chattopadhyay, and A.~Papandreou-Suppappola.
\newblock Monte carlo matching pursuit decomposition method for damage
  quantification in composite structures.
\newblock \emph{Journal of Intelligent Material Systems and Structures},
  20\penalty0 (6):\penalty0 647--658, 2009.

\bibitem[Mallat and Zhang(1993)]{Mallat_MPD_Dictionaries}
S.G. Mallat and Z.~Zhang.
\newblock Matching pursuits with time-frequency dictionaries.
\newblock \emph{IEEE Trans Signal Processing}, 41\penalty0 (12):\penalty0 3397
  --3415, 1993.

\bibitem[Das et~al.(2007)Das, Srivastava, and
  Chattopadhyay]{Das_et_al_SPIE_SVM}
S.~Das, A.~N. Srivastava, and A.~Chattopadhyay.
\newblock Classification of damage signatures in composite plates using
  one-class {SVM}s.
\newblock In \emph{Proc. IEEE Aerospace Conference}, pages 1 --19, 2007.

\bibitem[Azimipanah and ShahbazPanahi(2013)]{Azimipanah_compressive_imaging}
A.~Azimipanah and S.~ShahbazPanahi.
\newblock Experimental results of compressive sensing based imaging in
  ultrasonic non-destructive testing.
\newblock In \emph{2013 IEEE 5th International Workshop on Computational
  Advances in Multi-Sensor Adaptive Processing (CAMSAP)}, pages 336--339, 2013.

\bibitem[Sharma et~al.(2006)Sharma, Hanagud, and
  Ruzzene]{Sharma-et-al_Damage-Index_AIAA_2006}
V.~Sharma, S.~Hanagud, and M.~Ruzzene.
\newblock Damage index estimation in beams and plates using laser vibrometry.
\newblock \emph{AIAA Journal}, 44\penalty0 (4):\penalty0 919--923, 2006.

\bibitem[Michaels et~al.(2011)Michaels, Michaels, and
  Ruzzene]{Michaels_Ruzzene_Michaels_Ultrasonics_2010}
T.E. Michaels, J.E. Michaels, and M.~Ruzzene.
\newblock Frequency-wavenumber domain analysis of guided wavefields.
\newblock \emph{Ultrasonics}, 51\penalty0 (4):\penalty0 452 -- 466, 2011.

\bibitem[Basri and Chiu(2004)]{Chiu}
R.~Basri and W.~K. Chiu.
\newblock Numerical analysis on the interaction of guided {L}amb waves with a
  local elastic stiffness reduction in quasi-isotropic composite plate
  structures.
\newblock \emph{Composite Structures}, 66:\penalty0 87--99, 2004.

\bibitem[Alleyne and Cawley(1991)]{Alleyne}
D.~Alleyne and P.~Cawley.
\newblock A two-dimensional {F}ourier transform method for the measurement of
  propagating multimode signals.
\newblock \emph{Journal of the Acoustical Society of America}, 89:\penalty0
  1159--1168, 1991.

\bibitem[Ruzzene(2007)]{Ruzzene_SMS_2007}
M.~Ruzzene.
\newblock Frequency-wavenumber domain filtering for improved damage
  visualization.
\newblock \emph{Smart Materials and Structures}, 16\penalty0 (6):\penalty0
  2116, 2007.

\bibitem[Sohn et~al.(2011)Sohn, Dutta, Yang, DeSimio, Olson, and
  Swenson]{Sohn_Delamination_SMS_2011}
H.~Sohn, D.~Dutta, J.Y. Yang, M.~DeSimio, S.~Olson, and E.~Swenson.
\newblock Automated detection of delamination and disbond from wavefield images
  obtained using a scanning laser vibrometer.
\newblock \emph{Smart Materials and Structures}, 20\penalty0 (4):\penalty0
  045017, 2011.

\bibitem[An et~al.(2013)An, Park, and Sohn]{Sohn_Laser_Crack_SMS_2013}
Y.-K. An, B.~Park, and H.~Sohn.
\newblock Complete noncontact laser ultrasonic imaging for automated crack
  visualization in a plate.
\newblock \emph{Smart Materials and Structures}, 22\penalty0 (2):\penalty0
  025022, 2013.

\bibitem[Itti et~al.(1998)Itti, Koch, and Niebur]{Itti98}
L.~Itti, C.~Koch, and E.~Niebur.
\newblock A model of saliency-based visual attention for rapid scene analysis.
\newblock \emph{IEEE Transactions on Pattern Analysis and Machine
  Intelligence}, 20\penalty0 (11), 1998.

\bibitem[Itti and Koch(2001)]{Itti01}
L.~Itti and C.~Koch.
\newblock Computational modelling of visual attention.
\newblock \emph{Nature}, 2, March 2001.

\bibitem[Yan et~al.(2010)Yan, Zhu, Liu, and Liu]{Yan10}
J.~Yan, M.~Zhu, H.~Liu, and Y.~Liu.
\newblock Visual saliency detection via sparsity pursuit.
\newblock \emph{IEEE Signal Proc. Letters}, 17\penalty0 (8), 2010.

\bibitem[Shen and Wu(2012)]{Shen12}
X.~Shen and Y.~Wu.
\newblock A unified approach to salient object detection via low rank matrix
  recovery.
\newblock In \emph{Proc. Computer Vision and Pattern Recognition}, 2012.

\bibitem[Patcha and Park(2007)]{Patcha07}
A.~Patcha and J.~Park.
\newblock An overview of anomaly detection techniques: {E}xisting solutions and
  latest technological trends.
\newblock \emph{Computer Networks}, 51\penalty0 (12), 2007.

\bibitem[Chandola et~al.(2009)Chandola, Banerjee, and Kumar]{Chandola09}
V.~Chandola, A.~Banerjee, and V.~Kumar.
\newblock Anomaly detection: {A} survey.
\newblock \emph{ACM Computing Surveys}, 41\penalty0 (3), 2009.

\bibitem[Chen et~al.(2001)Chen, Donoho, and
  Saunders]{Chen_2001_ADB_588736_588850}
S.~S. Chen, D.~L. Donoho, and M.~A. Saunders.
\newblock Atomic decomposition by basis pursuit.
\newblock \emph{SIAM Rev.}, 43\penalty0 (1):\penalty0 129--159, 2001.

\bibitem[Candes et~al.(2006)Candes, Romberg, and Tao]{Candes1}
E.~Candes, J.~Romberg, and T.~Tao.
\newblock Robust uncertainty principles: {E}xact signal reconstruction from
  highly incomplete frequency information.
\newblock \emph{IEEE Transactions on Information Theory}, 52\penalty0
  (2):\penalty0 489--509, February 2006.

\bibitem[Donoho(2006)]{Donoho}
D.~Donoho.
\newblock Compressed sensing.
\newblock \emph{IEEE Transactions on Information Theory}, 52\penalty0
  (4):\penalty0 1289--1306, April 2006.

\bibitem[Cand\`{e}s and Tao(2006)]{Candes2}
E.~J. Cand\`{e}s and T.~Tao.
\newblock Near-optimal signal recovery from random projections: Universal
  encoding strategies?
\newblock \emph{IEEE Transactions on Information Theory}, 52\penalty0
  (12):\penalty0 5406--5425, December 2006.

\bibitem[Haupt and Nowak(2006)]{HauptRP}
J.~Haupt and R.~Nowak.
\newblock Signal reconstruction from noisy random projections.
\newblock \emph{{IEEE} Transactions on Information Theory}, 52\penalty0
  (9):\penalty0 4036--4048, September 2006.

\bibitem[Tropp(2004)]{tropp2004greed}
J.~A. Tropp.
\newblock Greed is good: {A}lgorithmic results for sparse approximation.
\newblock \emph{IEEE Transactions on Information Theory}, 50\penalty0
  (10):\penalty0 2231--2242, 2004.

\bibitem[Tropp(2006)]{tropp2006just}
J.~A. Tropp.
\newblock Just relax: {C}onvex programming methods for identifying sparse
  signals in noise.
\newblock \emph{IEEE Transactions on Information Theory}, 52\penalty0
  (3):\penalty0 1030--1051, 2006.

\bibitem[Bruckstein et~al.(2009)Bruckstein, Donoho, and Elad]{Bruckstein}
A.M. Bruckstein, D.L. Donoho, and M.~Elad.
\newblock From sparse solutions of systems of equations to sparse modeling of
  signals and images.
\newblock \emph{SIAM Rev.}, 51\penalty0 (1):\penalty0 34--81, 2009.

\bibitem[Gonella and Haupt(2013)]{Gonella_Haupt_IEEE_2013}
S.~Gonella and J.D. Haupt.
\newblock Automated defect localization via low rank plus outlier modeling of
  propagating wavefield data.
\newblock \emph{IEEE Transactions on Ultrasonics, Ferroelectrics and Frequency
  Control}, 60\penalty0 (12):\penalty0 2553--2565, 2013.

\bibitem[Olshausen and Field(1997)]{Olshausen97}
B.~A. Olshausen and D.~J. Field.
\newblock Sparse coding with an overcomplete basis set: {A} strategy employed
  by {V}1?
\newblock \emph{Vision Research}, 37:\penalty0 3311--3325, 1997.

\bibitem[Lewicki et~al.(1998)Lewicki, Sejnowski, and
  Hughes]{Lewicki98learningovercomplete}
M.~S. Lewicki, T.~J. Sejnowski, and H.~Hughes.
\newblock Learning overcomplete representations.
\newblock \emph{Neural Computation}, 12:\penalty0 337--365, 1998.

\bibitem[Kreutz-Delgado et~al.(2003)Kreutz-Delgado, Murray, Rao, Engan, Lee,
  and Sejnowski]{Kreutz03}
K.~Kreutz-Delgado, J.~F. Murray, B.~D. Rao, K.~Engan, T.-W. Lee, and T.~J.
  Sejnowski.
\newblock Dictionary learning algorithms for sparse representation.
\newblock \emph{Neural computation}, 15\penalty0 (2):\penalty0 349--396, 2003.

\bibitem[Aharon et~al.(2006)Aharon, Elad, and Bruckstein]{Aharon06}
M.~Aharon, M.~Elad, and A.~Bruckstein.
\newblock K-{SVD}: {A}n algorithm for designing overcomplete dictionaries for
  sparse representation.
\newblock \emph{IEEE Transactions on Signal Processing}, 54\penalty0
  (11):\penalty0 4311--4322, 2006.

\bibitem[Mairal et~al.(2010)Mairal, Bach, Ponce, and
  Sapiro]{Mairal_2010_OLM_1756006_1756008}
J.~Mairal, F.~Bach, J.~Ponce, and G.~Sapiro.
\newblock Online learning for matrix factorization and sparse coding.
\newblock \emph{J. Mach. Learn. Res.}, 11:\penalty0 19--60, 2010.

\bibitem[Mairal et~al.(2008)Mairal, Elad, and Sapiro]{mairal2008sparse}
J.~Mairal, M.~Elad, and G.~Sapiro.
\newblock Sparse representation for color image restoration.
\newblock \emph{IEEE Transactions on Image Processing}, 17\penalty0
  (1):\penalty0 53--69, 2008.

\bibitem[Peyr{\'e}(2009)]{peyre2009sparse}
G.~Peyr{\'e}.
\newblock Sparse modeling of textures.
\newblock \emph{Journal of Mathematical Imaging and Vision}, 34\penalty0
  (1):\penalty0 17--31, 2009.

\bibitem[Ramirez et~al.(2010)Ramirez, Sprechmann, and
  Sapiro]{ramirez2010classification}
I.~Ramirez, P.~Sprechmann, and G.~Sapiro.
\newblock Classification and clustering via dictionary learning with structured
  incoherence and shared features.
\newblock In \emph{Proc. IEEE Conf. on Computer Vision and Pattern
  Recognition}, pages 3501--3508, 2010.

\bibitem[Zubair et~al.(2013)Zubair, Yan, and Wang]{Zubair2013960}
S.~Zubair, F.~Yan, and W.~Wang.
\newblock Dictionary learning based sparse coefficients for audio
  classification with max and average pooling.
\newblock \emph{Digital Signal Processing}, 23\penalty0 (3):\penalty0 960 --
  970, 2013.

\bibitem[Rambhatla and Haupt(2013)]{Rambhatla:13}
S.~Rambhatla and J.~Haupt.
\newblock Semi-blind source separation via sparse representations and online
  dictionary learning.
\newblock In \emph{Proc. Asilomar Conf. on Signals, Systems, and Computers},
  Pacific Grove, CA, November 2013.

\bibitem[Li(2012)]{6469983}
Y.~Li.
\newblock Dictionary learning based multitask image restoration.
\newblock In \emph{Proc. Intl. Conf. on Image and Signal Processing}, pages
  364--368, 2012.

\bibitem[Mairal et~al.(2009)Mairal, Bach, Ponce, and Sapiro]{Mairal09}
J.~Mairal, F.~Bach, J.~Ponce, and G.~Sapiro.
\newblock Online dictionary learning for sparse coding.
\newblock In \emph{Proc. ICML}, 2009.

\end{thebibliography}

\end{document}